\documentclass[twocolumn,envcountsect]{svjour3}          
\smartqed  
\usepackage{graphicx}
\usepackage{mathptmx}      
\usepackage{hyperref}
\usepackage{amssymb}
\usepackage{algorithm}
\usepackage{algorithmic}
\usepackage{mathtools}

\usepackage{xfrac}

\usepackage[table]{xcolor}

\definecolor{light-gray}{gray}{0.9}

\journalname{Neuroinform}
\begin{document}

\title{Multi-Objective Cognitive Model: a supervised approach for multi-subject fMRI analysis}



\author{Muhammad Yousefnezhad \and Daoqiang Zhang}


\institute{The authors are with the College of Computer Science and Technology, Nanjing University of Aeronautics and Astronautics, Nanjing 211106, China.\\	
 \email{myousefnezhad@nuaa.edu.cn; dqzhang@nuaa.edu.cn}  
}

\date{Received: 27 Nov 2017 / Accepted: 12 Jul 2018}

\maketitle
\sloppy
\begin{abstract}
In order to decode human brain, Multivariate Pattern (MVP) classification generates cognitive models by using functional Magnetic Resonance Imaging (fMRI) datasets. As a standard pipeline in the MVP analysis, brain patterns in multi-subject fMRI dataset must be mapped to a shared space and then a classification model is generated by employing the mapped patterns. However, the MVP models may not provide stable performance on a new fMRI dataset because the standard pipeline uses disjoint steps for generating these models. Indeed, each step in the pipeline includes an objective function with independent optimization approach, where the best solution of each step may not be optimum for the next steps. For tackling the mentioned issue, this paper introduces Multi-Objective Cognitive Model (MOCM) that utilizes an integrated objective function for MVP analysis rather than just using those disjoint steps. For solving the integrated problem, we proposed a customized multi-objective optimization approach, where all possible solutions are firstly generated, and then our method ranks and selects the robust solutions as the final results. Empirical studies confirm that the proposed method can generate superior performance in comparison with other techniques.

\end{abstract}

\section{Introduction}
One of the primary goals in neuroscience is to understand how the neural activities in the human brain can be mapped to different cognitive tasks. Analyzing task-based functional Magnetic Resonance Imaging (fMRI) data is an interdisciplinary technique. Almost all supervised applications of fMRI analysis explicitly or implicitly employ Multivariate Pattern (MVP) algorithms for extracting and decoding brain patterns \cite{tony17b}. In practice, MVP analysis can be formulated as a classification problem and predict patterns of neural responses, which are generated by distinctive cognitive tasks \cite{tony17a}. In order to elaborate the brain mapping technique, imagine a subject watched two different categories of visual stimuli, including photos of cats and human faces, and we collected the neural activities in the form of fMRI dataset. Then, we have employed a subset of this data for training an MVP classification model in order to predict the categories (human face or cat) in the rest of stimuli (that are unseen in the training procedure). As the final product of an MVP analysis, decision surfaces are defined to distinguish the neural activities that belong to different categories of stimuli \cite{haxby14}. Decision surfaces can be used to understand mental diseases \cite{tony17a,tony17b}.

In practice, fMRI analysis is a challenging problem. Cognitive models generated by MVP methods must be validated by employing multi-subject fMRI images \cite{chen15,chen16,lorbert12,tony17a}. There are mainly two steps in an MVP standard pipeline, which must be applied to \emph{preprocessed} fMRI images to generate the cognitive models, i.e., functional alignment, classification analysis \cite{tony17b}. Further, each of these main steps can include some subtasks, such as applying feature selection before classification analysis \cite{chen15,tony17b}. Different human brains naturally generate distinctive patterns \cite{tony17a,haxby11}. The general assumption in the brain decoding is that the generated patterns are noisy `rotation' of a shared space \cite{chen15,haxby14,lorbert12,tony17a}. Functional alignment seeks this space for mapping the neural activities before generating the cognitive models \cite{haxby14}. As one of the common subtask in functional alignment, nonlinear kernel functions are employed to improve the performance of the alignment \cite{lorbert12}. The next step in the standard pipeline is employing binary classification methods, such as regularized Support Vector Machine (SVM) algorithm, for generating the cognitive models, i.e., decision surfaces  \cite{chen15,haxby14,lorbert12,mohr15,tony17a,tony17b}. Feature selection is a usual subtask, which must be applied before the classification analysis, to reduce the sparsity problem and increase the performance of the final model \cite{chen14}.

However, recent studies demonstrated that most of the models that generated by the standard pipeline cannot provide stable performance on the new fMRI datasets \cite{bennett09,chen16,eklund16,pauli16}. As discussed before, there are different steps (including the steps and subtasks) for MVP analysis. The main problem is that these steps employ disjoint objective functions, which are separately run to generate a cognitive model \cite{tony17b} and there is no unique solution for each of these objective functions \cite{chen15}. Therefore, an optimal result in one of these steps may not be optimum for the next steps. The problem gets worse when each step uses an independent optimization strategy, which cannot update the result of the previous steps based on the errors of current step to improve the performance of the MVP analysis. Indeed, this is a prevalent issue in the optimization problems, which is called dominant \cite{deb02,li16,li14,zitzler04}. In other words, if we have distinctive solutions, the ideal solution for each step not only must be an optimal solution for that step but also it must be optimum for all of the next steps. 

The main contributions of this paper are threefold: it firstly reformulates different steps of MVP pipeline as an integrated multi-objective problem, which is called Multi-Objective Cognitive Model (MOCM). We also introduce the novel concept of Intra-subject functional alignment, which can separately track the alignment error for each subject. Further, a customized optimization approach is developed for solving the integrated problem by incorporating the idea of non-dominated sorting into the multi-indicator algorithm. Indeed, non-dominated sorting seeks Pareto optimal solutions, and then indicators rank the robust solutions as the final results. 

In this paper, Section 2 briefly reviews some related works. Section 3 introduces the proposed method. Section 4 reports empirical studies. Finally, Section 5 presents conclusion and pointed out some future studies.
\section{Background}
Since task-based fMRI datasets can provide better spatial resolution in comparison with other modalities, most of the previous studies employed fMRI datasets in order to study human brains \cite{haxby14}. A crucial step in fMRI analysis is creating a model that is generalized across subjects \cite{chen15,chen14,chen16,hanke14,haxby11,xu12,tony17a}. In other words, utilizing multi-subject fMRI data is necessary to validate the generated results across subjects \cite{haxby11,tony17a}. However, functional neural images require precise alignment for boosting the performance of the final model \cite{chen14,haxby11,tony17a}. In practice, they are two primary alignment approaches, including anatomical alignment and functional alignment, that must work in unison \cite{haxby11,xu12}. Indeed, anatomical alignment is a classical technique for preprocessing fMRI images. However, the performance of the anatomical alignment is limited based on the location, shape, and size of the functional loci \cite{rademacher93,watson93}. By contrast, functional alignment does not suffer the mentioned issues in the anatomical alignment \cite{haxby11}.

Hyperalignment (HA), as the most prevalent approaches for applying functional alignment, is an `anatomy free' technique that can be written as a Canonical Correlation Analysis (CCA) problem \cite{haxby14,haxby11,tony17a}. As discussed before, HA seeks a shared neural representation across subjects. The performance of MVP analysis by using functional alignment is significantly increased \cite{haxby14,haxby11}. Xu et al. proposed the Regularized Hyperalignment (RHA), which uses an EM algorithm to find the optimum template parameters iteratively \cite{xu12}. Lorbert et al. developed Kernel Hyperalignment (KHA) as a nonlinear extension of HA method \cite{lorbert12}. Chen et al. introduced a two-phase method for functional alignment, which is called Singular Value Decomposition Hyperalignment (SVDHA). The neural activities are mapped in SVDHA to a low-dimensional space by using SVD, and then the mapped features are aligned by using HA techniques \cite{chen14}. Yousefnezhad et al. introduced Local Discriminant Hyperalignment (LDHA) as the first supervised HA method for MVP analysis \cite{tony17a}.

MVP techniques utilize classification algorithms for predicting a new subject's neural activities. Cox et al. used both linear and nonlinear SVM algorithms for the human brain decoding \cite{cox03}. Norman et al. illustrated SVM superior performance to Gaussian Naive Bayes classifiers \cite{norman06}. Carroll et al. developed a new cognitive model by employing the Elastic Net \cite{zou05} to predict and interpret the distributed neural responses with sparse models \cite{carroll09}. Mohr et al. analyzed different classification techniques, i.e., the first norm regularized SVM \cite{bradley98}, the second norm regularized SVM \cite{cortes95}, the Elastic Net \cite{zou05}, and the Graph Net \cite{grosenick13}, to predict distinctive neural activities in the human brain \cite{mohr15}. They figured out the first norm regularized SVM can rapidly improve the classification performance in fMRI analysis \cite{mohr15,tony17b}. Osher et al. developed a network-based method by employing the human brain's anatomical features in order to classify distinctive neural responses \cite{osher15}. Yousefnezhad et al. proposed two new ensemble learning approaches by utilizing weighted AdaBoost \cite{tony16}, and Bagging \cite{tony17b}.

Recent studies demonstrated that most of the generated models by the standard pipeline cannot provide stable performances on new fMRI datasets \cite{bennett09,chen16,eklund16,pauli16}. Some studies illustrated that results of General Linear Models (GLM) that are generated by different software packages (AFNI \cite{cox96}, FSL \cite{jenkinson12}, and SPM \cite{penny11}) on a specific problem can be highly unstable \cite{pauli16}. Since these linear models are utilized in most of fMRI analysis (i.e., MVP classification), unstable models rapidly decrease robustness of the final results \cite{bennett09,cai16,pauli16}. Eklund et al. proved that the cognitive models generated by rest-mode fMRI datasets for spatial extent could significantly increase inflated false-positive rates \cite{eklund16}. Chen et al. developed Convolutional Autoencoder (CAE) method for improving the stability of functional alignment to analyze the whole brain neural activities \cite{chen16}. Indeed, CAE employed Shared Response Model (SRM) \cite{chen15} for functional alignment as well as the standard searchlight analysis \cite{guntupalli16} for improving the stability of the generated cognitive model \cite{chen15}.

There are a few studies that used multi-objective optimization \cite{carroll09,kao09,kao12}. Indeed, these approaches formulate different steps belonging to fMRI analysis by using multiple objective functions rather than using single objectives for each section. Then, these methods seek optimal solutions for all of the objective functions simultaneously. Here, we may seek multiple optimal solutions for a specific problem, where they must be ranked in order to find the best final result. There are two approaches for ranking better solutions that must work in unison, i.e., non-dominated sorting \cite{deb02} and multi-indicator algorithm \cite{li16}. While non-dominated sorting seeks all possible solutions, indicators rank the robust solutions in each Pareto frontier as the final results. In fMRI studies, Kao proposed a multi-objective approach for estimating a general linear model between the design matrix and the neural activities \cite{kao09}. Conroy et al. develop a multi-objective optimization for selecting models in fMRI analysis, where they provided a principled method to take into account both classification accuracy and stability \cite{conroy13}. In another study, Kao et al. utilized  a modified version of Non-dominated Sorting Genetic Algorithm (NSGA-II) for generating the linear model between the fMRI responses and task-related events \cite{kao12}. Ma et al. developed a  multi-objective MVP technique by using Hierarchical Heterogeneous Particle Swarm Optimization (HHPSO), where the classification problem is formulated as a binary SVM, and then HHPSO seeks optimal solutions \cite{ma16}.

\begin{figure}[!t]
	\begin{center}
		\includegraphics[width=0.48\textwidth,height=0.98\linewidth]{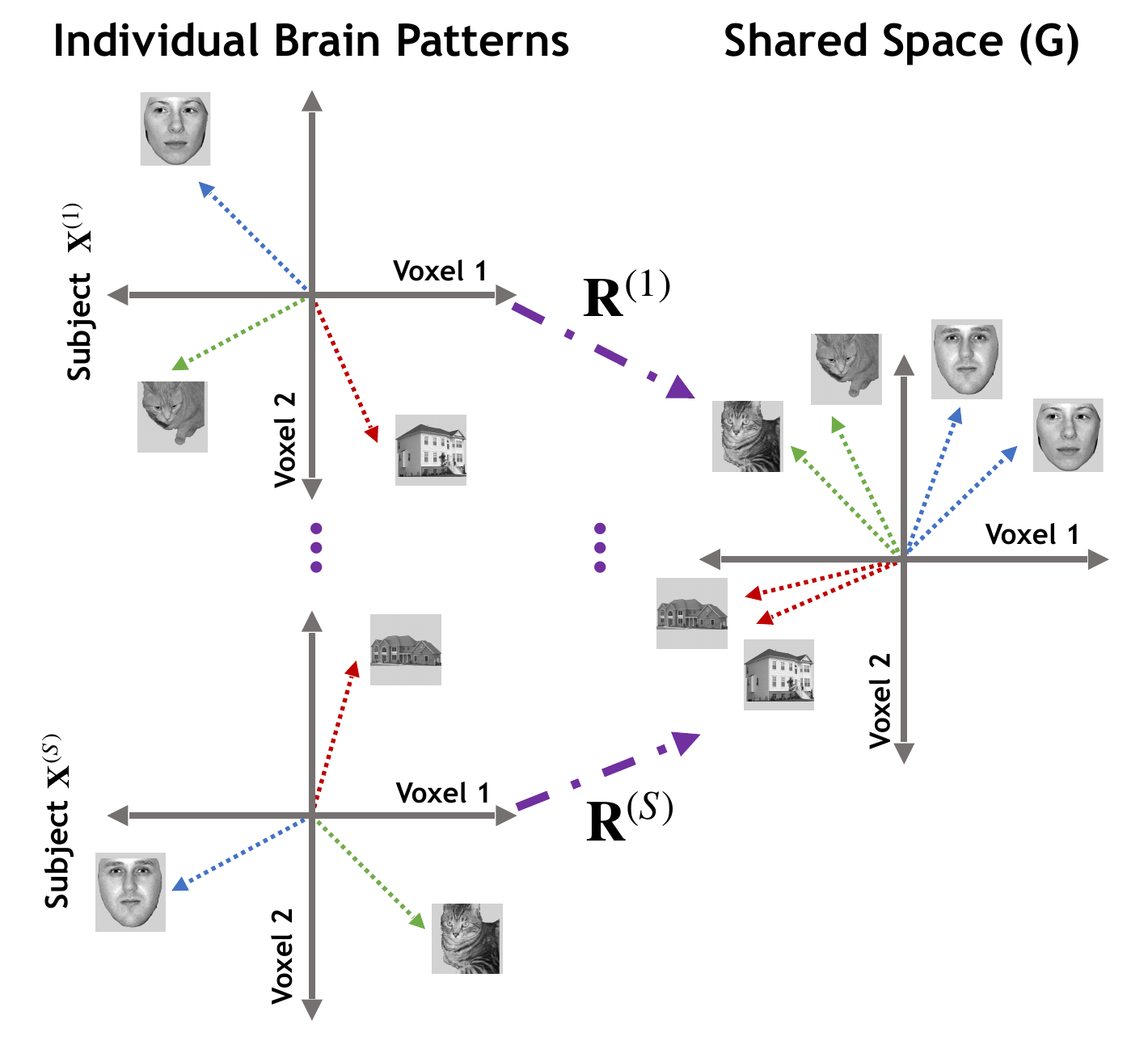}
		\caption{An example of functional alignment in a two-voxel representation space. Here, the neural activities are depicted by vectors with different colors, and each color (i.e., blue, green, and red) represents a specific category of visual stimuli. Further, $R^{(\ell)}$ denotes the mapping from the original features to the shared space \cite{haxby11,lorbert12,xu12,tony17a}.}
		\label{fig:FuncAlign}
	\end{center}
	\vskip -0.2in
\end{figure}

\section{The Proposed Method}
As preprocessed fMRI dataset, $\mathbf{F}^{(i)} \in \mathbb{R}^{T \times V_{org}}\text{, } i=1\text{:}S$ is defined, where $S$ denotes the number of subjects, $V_{org}$ is the number of voxels in the original space, $T$ denotes the number of time points in units of Time of Repetitions (TRs). This paper assumes that the neural activities of each subject are column-wise standardized, i.e., $\mathbf{F}^{(i)} \sim {N}(0,1)$. We can also consider this condition as a preprocessing step if the original data is not standardized. A linear model then can be formulated for each subject as follows:
\begin{equation}\label{eq:LinearModel}
	\mathbf{F}^{(i)} = \mathbf{D}^{(i)}\mathbf{\beta}^{(i)} + \mathbf{\varepsilon}^{(i)}\text{, }i=1\text{:}S 
\end{equation}
where $\mathbf{D}^{(i)} \in \mathbb{R}^{T\times C}$ denotes the design matrix, $\varepsilon^{(i)}$ is the error of estimation, ${\beta^{(i)}} \in \mathbb{R}^{C\times V_{org}}$ is the sets of regressors. In addition, $C$ denotes the number of stimulus categories in the experiment \cite{tony16,tony17b}. Design matrix can be generated by convolution of time samples (or onsets: $\mathbf{\tau}^{(i)} \in \mathbb{R}^{T\times C}$) and the Hemodynamic Response Function (HRF) signal ($\mathcal{H}$), i.e., $\mathbf{D}^{(i)} = \mathbf{\tau}^{(i)}*\mathcal{H}$ \cite{tony17b}. Here, time synchronized stimulus ensures temporal alignment, including the $m\text{-}th$ time point for all of the subjects represents the same simulation \cite{lorbert12,xu12,tony17a}. In order to estimate ${\beta^{(i)}}$, the first objective function is defined for tracking error ($\varepsilon^{(i)}$) in \eqref{eq:LinearModel} as follows:
\begin{equation}
	\begin{split}\label{eq:ThetaLiearModel}
		\theta_1 = \frac{1}{S}\sum_{i=1}^{S}\mathbf{F}^{(i)} - \mathbf{D}^{(i)}\mathbf{\beta}^{(i)}
	\end{split}
\end{equation}
By considering \eqref{eq:ThetaLiearModel}, the stimuli in the training-set are considered time synchronized, i.e., each time point for all subjects illustrates the same simulation \cite{lorbert12,xu12}. Consequently, the class labels for the training-set are defined by $\mathbf{Y}=\Big\{{y}_m \Big\}\text{, }{y}_m \in \{-1,+1\}\text{, } m=1\text{:}T$. In order to generalize the proposed method, a mapping function can be defined as follows:

\begin{equation}
	\begin{split}\label{eq:KernelFunc}
		\mathbf{X^{(\ell)}} \in \mathbb{R}^{T \times V} = {\Phi}\Big(\mathbf{D^{(\ell)}}\mathbf{{\beta^{(\ell)}}}\Big)
	\end{split}
\end{equation}
where ${\Phi}:\mathbb{R}^{T\times V_{org}}\rightarrow \mathbb{R}^{T\times V}$ can be considered for two different applications. It can be any kernel function \cite{lorbert12} that maps the voxels from original nonlinear space to a linear embedded space. Further, this function can be any feature selection/ranking function \cite{chen15,chen14,gucclu15}. In order to employ the original data, this function can be considered as a linear mapping, where ${\Phi}(\mathbf{x})=\mathbf{x}$. We will analyze different applications of this function in the experiments section.

The next step is functional alignment. As mentioned before, the general assumption in the brain decoding is that the generated patterns in each brain are noisy `rotation' of a shared space \cite{haxby14,tony17a,chen15,lorbert12}. Figure \ref{fig:FuncAlign} illustrates an example for functional alignment in a two-voxel representation space. As depicted in this figure, Hyperalignment (HA) seeks a shared space by using the training-set, where the correlations between different stimuli are minimized. By considering \eqref{eq:ThetaLiearModel}, functional alignment can be defined as follows \cite{tony17a}:
\begin{equation}
	\begin{split}\label{eq:HA}
		\theta_2 =\frac{1}{S} \sum_{\substack{i = 1\\j=i+1}}^{S}\big\|  \mathbf{X}^{(i)}\mathbf{R}^{(i)} -  \mathbf{X}^{(j)}\mathbf{R}^{(j)} \big\|^2_F\qquad \\
		\text{s.t.}\qquad\big(\mathbf{X}^{(\ell)}\mathbf{R}^{(\ell)}\big)^\top\mathbf{X}^{(\ell)}\mathbf{R}^{(\ell)}=\mathbf{I}\text{, }\qquad \ell=1\text{:}S
	\end{split}
\end{equation}
where $\mathbf{I}$ is the identity matrix, and $\mathbf{R}^{(\ell)}\in \mathbb{R}^{V \times V}$ denotes the mapping that must be calculated for each subject. Here, voxel correlation map $(\mathbf{X}^{(i)})^\top \mathbf{X}^{(j)}$, $i,j=1\text{:}S$ in the most of fMRI studies is not full rank because the number of voxels is significantly more than TRs \cite{chen14,lorbert12,xu12,tony17a}. Since \eqref{eq:HA} must be calculated for any new subject in the testing-phase, it is not computationally efficient.  
\begin{lemma}
	\emph{The equation \eqref{eq:HA} is equivalent to:}
	\begin{equation}
		\begin{split}\label{eq:CCA}
			\theta_2 = \frac{1}{S} \sum_{i=1}^{S} \big\| \mathbf{G} - \mathbf{X}^{(i)}\mathbf{R}^{(i)}\big\|^2_F \qquad \\ \text{s.t.}\qquad\big(\mathbf{X}^{(\ell)}\mathbf{R}^{(\ell)}\big)^\top\mathbf{X}^{(\ell)}\mathbf{R}^{(\ell)}=\mathbf{I}\text{, }\qquad \ell=1\text{:}S
		\end{split}
	\end{equation}
	\emph{where $\mathbf{G} \in \mathbb{R}^{T\times V}$ is the HA shared space:}
	\begin{equation}
		\begin{split}\label{eq:G}
			\mathbf{G} = \frac{1}{S} \sum_{j=1}^{S} \mathbf{X}^{(j)}\mathbf{R}^{(j)}
		\end{split}
	\end{equation}	
	\emph{Proof.} In a nutshell, both \eqref{eq:HA} and \eqref{eq:CCA} can be reformulated as $S^2\emph{tr}\big(\mathbf{G}^\intercal\mathbf{G}\big) - \bigg(S\sum_{i=1}^{S}\emph{tr}\Big(\big(\mathbf{X}^{(i)}\mathbf{R}^{(i)}\big)^\intercal\mathbf{X}^{(i)}\mathbf{R}^{(i)}\Big)\bigg)$, where $\emph{tr()}$ denotes the trace function. Please see \cite{tony17a,lorbert12} for details.
\end{lemma}

$\mathbf{G}$ is called the HA shared space, which can be used for functional alignment in the testing-phase \cite{tony17a,xu12,haxby11,chen14}. 

\begin{remark}\label{rm:InterSubjIssue}\emph{The angle of rotation for all stimuli in each subject after mapping must be equal. This paper defines Intra-Subject Evaluation (ISE) as follows for calculating the angle of rotation for each category of stimuli:
		\begin{equation}\label{eq:ISE}
			\emph{ISE}(\mathbf{x},\mathbf{g})= \frac{\mathbf{x^\top g}}{\|\mathbf{x}\|\|\mathbf{g}\|}
		\end{equation}
		where the vectors $\mathbf{x}\in \mathbb{R}^{V}$ and $\mathbf{g}\in \mathbb{R}^{V}$ respectively denote the neural activities in a specific time point before and after mapping.} 
\end{remark}
By considering \eqref{eq:ISE}, the error of rotation for all subject is calculated as follows:
\begin{equation}\label{eq:IntraSubjError}
	\begin{split}
		\theta_3 = \frac{1}{S} \sum_{\ell=1}^{S}\sum_{\substack{m=1\\n=m+1}}^{C}\Big\| \text{ISE}\big(\mathbf{x}^{(\ell)}_{m.},\mathbf{g}^{(\ell)}_{m.}\big) - \text{ISE}\big(\mathbf{x}^{(\ell)}_{n.},\mathbf{g}^{(\ell)}_{n.}\big)\Big\|^2_F
	\end{split}
\end{equation}
where row vector $\mathbf{x}^{(\ell)}_{m.} \in \mathbb{R}^{V}$ denotes the neural activities (all voxels) belong to $m\text{-}th$ time point, i.e., $\mathbf{x}^{(\ell)}_{m.} = \{x^{(\ell)}_{mn} | x^{(\ell)}_{mn} \in \mathbf{X}^{(\ell)}, n = 1:V\}$. Further, we have the same notation for the shared space, $\mathbf{g}^{(\ell)}_{m.}\in \mathbb{R}^{V} = \{g^{(\ell)}_{mn} | g^{(\ell)}_{mn} \in \mathbf{G}^{(\ell)}, n = 1:V\}$. 

\begin{remark}\emph{
		As mentioned before, classification algorithms are employed in MVP analysis for generating the cognitive model. While we can use any algorithm for training a cognitive model, this paper employs L1 regularized SVM \cite{bradley98} that is utilized in \cite{mohr15} as the best algorithm for fMRI analysis.}
\end{remark}
As the next step, a classification model is defined as follows:
\begin{equation}\label{eq:Clf}
	\theta_4 = \frac{\alpha}{S} \sum_{i=1}^{S} {\max}\bigg(0, \mathbf{1}_{T} - \Big( \text{diag}(\mathbf{Y})\mathbf{X}^{(i)}\mathbf{R}^{(i)}\mathbf{W}\Big)\bigg)^2 + \| \mathbf{W} \|_1
\end{equation}
where the constraint $\alpha > 0$ is the SVM parameter, $diag$ function create a square diagonal matrix from the class label vector $\mathbf{Y}$, $\mathbf{1}_{T} \in \mathbb{R}^{T}$ is ones vector, $\| . \|$ denotes the L1 norm, $\mathbf{W} \in \mathbb{R}^{T}$ is the decision surfaces for our cognitive model.

Training-phase for MOCM can be denoted by using following objective function:
\begin{equation}\label{eq:ClfTrain}
	\begin{split}
		\min_{\mathbf{p}_{train}}{\Theta}_{train}\qquad\qquad\qquad\\ \text{s.t.}\quad{\Theta}_{train}={K}_{train}\big(\mathbf{F},\mathbf{\tau};\mathbf{p}_{train}\big)
	\end{split}
\end{equation}
where the vector ${\Theta}_{train} = \big[\theta_1, \theta_2, \theta_3, \theta_4 \big]$ is the training error, the fMRI time series $\mathbf{F} = \{\mathbf{F}^{(i)},  i=1\text{:}S\}$ and its corresponding onsets $\mathbf{\tau} = \{\mathbf{\tau}^{(i)}, i=1\text{:}S\}$ are considered as the training-set, and the training parameters are defined by the vector $\mathbf{p}_{train} = \big[\beta^{(\ell)}, \mathbf{R}^{(\ell)}, \mathbf{W}\big]\text{, }\ell=1\text{:}S$. Here, ${K}_{train}$ respectively employs \eqref{eq:ThetaLiearModel}, \eqref{eq:HA}, \eqref{eq:IntraSubjError}, \eqref{eq:Clf} in order to estimate $\theta_1, \theta_2, \theta_3,$ and $\theta_4$. In testing-phase, the following objective function is used:
\begin{equation}\label{eq:ClfTest}
	\begin{split}
		\min_{\mathbf{p}_{test}}{\Theta}_{test}\qquad\qquad\qquad\\ \text{s.t.}\quad{\Theta}_{test}={K}_{test}\big(\mathbf{\widehat{F}},\mathbf{\widehat{\tau}},\mathbf{G};\mathbf{p}_{test}\big)
	\end{split}
\end{equation}
where the vector ${\Theta}_{test} = \big[\theta_1, \theta_2, \theta_3 \big]$ is the error of testing-phase, the fMRI time series $\mathbf{\widehat{F}} = \{\mathbf{\widehat{F}}^{(i)},  i=1\text{:}\widehat{S}\}$ and its corresponding onsets $\mathbf{\widehat{\tau}}^{(i)} = \{\mathbf{\widehat{\tau}}^{(i)},  i=1\text{:}\widehat{S}\}$ denote the testing-set, $\widehat{S}$ is the number of subjects in the testing-set, $\mathbf{G}$ denotes the shared space that is calculated in the training-phase, and the testing parameters are defined by the vector $\mathbf{p}_{test} = \big[\widehat{\beta}^{(\ell)}, \mathbf{\widehat{R}}^{(\ell)}\big]\text{, }\\\ell=1\text{:}S$. Here, ${K}_{test}$ respectively uses \eqref{eq:ThetaLiearModel}, \eqref{eq:CCA}, \eqref{eq:IntraSubjError} for estimating $\theta_1, \theta_2, \text{ and }\theta_3$. Furthermore, the final prediction can be generated by $\widehat{\mathbf{Y}}^{(i)} \in \mathbb{R}^{T}= \mathbf{1}_{T} - \Big(\mathbf{\widehat{X}}^{(i)}\mathbf{\widehat{R}}^{(i)}\mathbf{W}\Big)$ for all new subjects ($i=1\text{:}\widehat{S}$), where $\mathbf{W}$ denotes the decision surfaces that is calculated in the training-phase.
\begin{algorithm}[!t]
	\caption{Multi-Objective Cognitive Model (MOCM)}
	\label{alg:MOCM}
	\begin{algorithmic}
		\STATE {\bfseries Input:} ${K}(\mathbf{p})$ for training-phase or testing-phase,\\ 
		$O$ as population size, $MaxIt$ as maximum iterations,
		$MaxSame$ as maximum iterations with the same optimal result.
		\STATE {\bfseries Output:} An optimal solution $p_{opt}$.\\
		\STATE {\bfseries Method:}\\
		01. Random initial $\mathbf{P}^{(0)}$ as a set of the 1st $O$ solutions.\\ 
		02. $i = 0$, $j = 0$.\\
		03. \textbf{While} $(i < MaxIt)$ \text{ and } $(j < MaxSame)$  \textbf{Do}\\ 
		04. \quad$\mathbf{Q}^{(i)} = \emptyset$.\\
		05. \quad\textbf{Repeat} $O$-times \textbf{Do}\\
		06. \quad\quad Randomly choose $\mathbf{p}_1,\mathbf{p}_2 \in \mathbf{P}^{(i)}$.\\
		07. \quad\quad Create offspring $\mathbf{q} = (\mathbf{p}_1 + \mathbf{p}_2)/2$.\\
		08. \quad\quad Update $\mathbf{Q}^{(i)} = \mathbf{Q}^{(i)} \cup \{\mathbf{q}\}$.\\
		09. \quad\textbf{End Repeat}\\
		10. \quad Random initial $\mathbf{E}^{(i)}$ with size $O$.\\
		11. \quad Generate new population: $\mathbf{U}^{(i)} = \mathbf{P}^{(i)} \cap \mathbf{Q}^{(i)} \cap \mathbf{E}^{(i)}$.\\
		12. \quad$\mathbf{P}^{(i+1)}$ = \textbf{SORT}$\Big({K}(\mathbf{p})$, $\mathbf{U}^{(i)}$, $O\Big)$.\\ 
		13. \quad$\mathbf{p}_{opt}^{(i)}$ as the first sorted solution in $\mathbf{P}^{(i+1)}$.\\
		14. \quad\textbf{If} $\big(K(\mathbf{p}_{opt}^{(i)}) == K(\mathbf{p}_{opt}^{(i-1)})\big)$\\
		15. \qquad$j = j+1$.\\
		16. \quad\textbf{Else}\\
		17. \qquad$j=0$.\\
		18. \quad\textbf{End If}\\
		19. \quad$i = i+1$.\\
		20. \textbf{End While}\\
		21. Return $\mathbf{p}_{opt}^{(i)}$.\\
	\end{algorithmic}
\end{algorithm}

\subsection{Optimization}
In the previous section, we introduced an integrated multi-objective function in order to apply supervised fMRI analysis. This section presents a customized multi-objective optimization approach for finding optimal solutions for both the training-phase and testing-phase. For simplicity, a generalized objective function is considered as follows:
\begin{equation}
	\begin{split}
		{\Theta} \leftarrow {K}(\mathbf{p})
	\end{split}
\end{equation}
where the function ${K}$ and parameters $\mathbf{p}$ can be calculated by \eqref{eq:ClfTrain} for training-phase, and \eqref{eq:ClfTest} is used for the testing-phase. Algorithm \ref{alg:MOCM} depicts a general template as the optimization approach for MOCM. In this algorithm, $O$ is the population size, $MaxIt$ denotes the maximum number of iterations, and $MaxSame$ is the maximum number of iterations with the same optimal solution. This algorithm firstly considers a set of $O$ random solutions ($\mathbf{P}^{(0)}$) for the first iteration. Further, we create three different sets of solutions (with size $O$) in each iteration in order to generate a new population for the next iteration. As the first set, $\mathbf{P}^{(i)}$ is the best $O$ solutions that are generated in the previous step. As the second set, we create $O$ new offsprings by averaging randomly selected parents from $\mathbf{P}^{(i)}$. Indeed, this set tries to seek better solutions by combining the previous best solutions. As the last set, we create $O$ new random solutions ($\mathbf{E}^{(i)}$) in order to increase the diversity of possible solutions. In fact, $\mathbf{E}^{(i)}$ can rapidly reduce the chance of the local optimum issue. These sets are combined as a new population ($\mathbf{U}^{(i)}$) with size $3O$, and then the $\textbf{SORT()}$ function select the first $O$ optimal solutions for the next step. Further, the first sorted solution ($\mathbf{p}_{opt}^{(i)}$) from $\mathbf{P}^{(i+1)}$ is considered as the best solution for $i\text{-}th$ iteration. As the finishing condition, the algorithm repeats $MaxIt$-times the optimization procedure unless the best solutions for $MaxSame$-times will be the same.

\begin{algorithm}[!t]
	\caption{The $\textbf{SORT}$ function}
	\label{alg:SORT}
	\begin{algorithmic}
		\STATE {\bfseries Input:} Objective function ${K}(\mathbf{p})$, A set of solutions $\mathbf{U}$,\\ \quad Population size $O$\\
		\STATE {\bfseries Output:} $\mathbf{P}_{opt}$ as the first $O$ optimal solutions. \\
		\STATE {\bfseries Method:}\\
		01. \textbf{For Each} $\mathbf{p} \in \mathbf{U}$\\
		02. \quad$\Delta_p = \emptyset$, $n_p = 0$.\\
		03. \quad\textbf{For Each} $(\mathbf{q} \in \mathbf{U})$ \textbf{and} $(\mathbf{q}\neq \mathbf{p})$\\
		04. \quad\quad$\Theta_p \leftarrow {K}(\mathbf{p})$, $\Theta_q \leftarrow {K}(\mathbf{q})$\\
		05. \quad\quad\textbf{If} $({\Theta}_p \prec {\Theta}_q)$ \\
		06. \quad\quad\qquad${\Delta}_p = {\Delta}_p \cup \{\mathbf{q}\}$.\\
		07. \quad\quad\textbf{Elsif} $\big({\Theta}_q \prec {\Theta}_p\big)$\\ 
		08. \quad\quad\qquad$n_p = n_p + 1$.\\
		09. \quad\quad\textbf{End If}\\
		10. \quad\textbf{End For}\\
		11. \quad\textbf{If} $n_p = 0$ \textbf{Then}\\
		12. \quad\quad${\Omega}^{(1)} = {\Omega}^{(1)} \cup \{\mathbf{p}\}.$\\
		13. \quad\textbf{End If}\\
		14. \textbf{End For}\\
		15. $j = 1$, $\mathbf{P}_{opt} = \emptyset$.\\
		16. \textbf{While} $({\Omega}^{(j)}\neq\emptyset)$ \textbf{and} $(| \mathbf{P}_{opt} | < O)$\\
		17. \quad${\Omega}^{(j+1)} = \emptyset$.\\
		18. \quad $\mathbf{I1}$-evaluation: ${a}_p = \mathbf{I1}\big(\mathbf{p},{\Omega}^{(j)}\big) \text{ for all } \mathbf{p} \in {\Omega}^{(j)}$. \\
		19. \quad $\mathbf{I2}$-evaluation: ${b}_p = \mathbf{I2}\big(\mathbf{p},{\Omega}^{(j)}\big) \text{ for all } \mathbf{p} \in {\Omega}^{(j)}$. \\		
		20. \quad Order ${\Omega}^{(j)}$ (solutions with lowest $\max$(${a}_p$, ${b}_p$) in top).\\
		21. \quad\textbf{For Each} $(\mathbf{p} \in {\Omega}^{(j)})$\\
		22. \quad\quad$\mathbf{P}_{opt} = \mathbf{P}_{opt} \cup \{\mathbf{p}\}$.\\
		23. \quad\quad\textbf{For Each} $(\mathbf{q} \in {\Delta}_p)$\\
		24. \quad\quad\quad$n_q = n_q - 1$.\\
		25. \quad\quad\quad\textbf{If} $(n_q = 0)$\\
		26. \quad\quad\quad\quad${\Omega}^{(j+1)} = {\Omega}^{(j+1)} \cup \{\mathbf{q}\}$.\\
		27. \quad\quad\quad\textbf{End If}\\
		28. \quad\quad\textbf{End For}\\
		29. \quad\textbf{End For}\\
		30. \quad$j = j + 1$.\\
		31. \textbf{End While} 
	\end{algorithmic}
\end{algorithm}

The key point in Algorithm \ref{alg:MOCM} is the $\textbf{SORT()}$ function. Algorithm \ref{alg:SORT} illustrates this function. As mentioned before, the optimization approach for MOCM is developed by incorporating the idea of non-dominated sorting \cite{deb02} into the multi-indicator algorithm \cite{li16}. As the first step, the solutions in $\mathbf{U}$ are ranked based on the concept of domination. 
\begin{remark}\label{rm:Dominant}\emph{
		The best solution ($\mathbf{p}_{opt}$) must dominate ($\prec$) \cite{deb02,li16} all possible solutions in $\mathbf{U}$. In other words, the estimation of the optimal result (${\Theta}_{opt} \leftarrow {K}(\mathbf{p}_{opt})$) must satisfy the following conditions in comparison with all possible estimations (${\Theta}_q \leftarrow {K}(\mathbf{q}) \text{ for all } q \in \mathbf{U}$):}
	\begin{equation}
		\begin{split}
			\forall \theta_{opt}^{(j)} \in {\Theta}_{opt} \text{, } \theta_{q}^{(\ell)} \in {\Theta}_q \implies \theta_{opt}^{(j)} \leq \theta_{q}^{(\ell)}\\
			\exists \theta_{opt}^{(j)} \in {\Theta}_{opt} \text{, } \theta_{q}^{(\ell)} \in {\Theta}_q \implies \theta_{opt}^{(j)} < \theta_{q}^{(\ell)}
		\end{split}
	\end{equation}
\end{remark}

In order to apply non-dominated sorting, Algorithm \ref{alg:SORT} firstly generates two criteria for ranking all possible solutions ($\mathbf{U}$), i.e., the scale $n_p$ and the matrix ${\Delta}_{p}$ for each solution.  The scale $n_p$ counts the number of solutions that can dominate the solution $\mathbf{p}$, and the matrix ${\Delta}_{p}$ denotes the set of solutions that are dominated by the solution $\mathbf{p}$. Further, the set of the first Front (${\Omega}^{(1)}$) can be defined by the solutions that are not dominated by any solution ($\forall p\in{\Omega}^{(1)} \implies n_p=0$).

As the second step, Algorithm \ref{alg:SORT} must create $O$ optimal sorted solutions. As mentioned before, this paper uses the multi-indicator algorithm for evaluating error rates in the possible results. Indeed, these indicators can evaluate the robustness of the generated results. This paper employs two effective indicators, i.e., $\mathbf{I1}$ \cite{li16,zitzler04} and $\mathbf{I2}$ \cite{li16}. As the first indicator, $\mathbf{I1}$ is defined as follows \cite{li16,zitzler04}:
\begin{equation}\label{eq:I1}
	\mathbf{I1}(\mathbf{q},\mathbf{P})=\sum_{\substack{\mathbf{p}\in\mathbf{P}\\\mathbf{p}\neq\mathbf{q}}}\exp(\frac{-1}{0.05}\mathbf{I_{\epsilon+}}(\mathbf{q},\mathbf{p}))
\end{equation}
where $\mathbf{I_{\epsilon+}}(\mathbf{p},\mathbf{q})$ is denoted as follows \cite{zitzler04}:
\begin{equation}\label{eq:Ieps}
	\begin{split}
		\mathbf{I_{\epsilon+}}(\mathbf{p},\mathbf{q}) = \underset{\epsilon}{\min}\big(\theta_\ell-\epsilon \leq \widehat{\theta_\ell}\big)\text{,\qquad} \substack{\forall \theta_\ell \in {\Theta_p} \leftarrow {K}(\mathbf{p}) \\\forall \widehat{\theta_\ell} \in {\Theta_q} \leftarrow {K}(\mathbf{q})}
	\end{split}
\end{equation}
Further, indicator $\mathbf{I2}$ is defined as follows \cite{li16}:
\begin{equation}\label{eq:I2}
	\mathbf{I2}(\mathbf{q},\mathbf{P})=\underset{\substack{\mathbf{p}\in \mathbf{P}\\\mathbf{p}\text{ precedes }\mathbf{q}}}{\min} \big(\mathbf{I_{SDE}}(\mathbf{q},\mathbf{p})\big)
\end{equation}
where $\mathbf{p}$ precedes $\mathbf{q}$ means that the position (the original index) of $\mathbf{p}$ in the population $\mathbf{P}$ is smaller than the position $\mathbf{q}$ \cite{li16}. In addition, $\mathbf{I_{SDE}}$ is calculated as follows \cite{li14}:
\begin{equation}\label{eq:ISDE}
	\mathbf{I_{SDE}}(\mathbf{p},\mathbf{q})=\Big(\sum_{\ell}\text{sd}\big(\theta_\ell,\widehat{\theta_\ell}\big)^2 \Big)^{\frac{1}{2}}\text{,\qquad}\substack{\forall \theta_\ell \in {\Theta_p} \leftarrow {K}(\mathbf{p}) \\\forall \widehat{\theta_\ell} \in {\Theta_q} \leftarrow {K}(\mathbf{q})}
\end{equation}
\begin{equation}
	\begin{split}
		\text{sd}(\theta, \widehat{\theta})=\begin{cases}
			\theta - \widehat{\theta}       & \quad \text{if } \theta < \widehat{\theta}\\
			0  & \quad otherwise.\\
		\end{cases}
	\end{split}
\end{equation}
In order to select the $O$ optimal solutions, the set of $j\text{-}th$ Front solutions (${\Omega}^{(j)}$) will be evaluated by $\mathbf{I1}$ and $\mathbf{I2}$. Then, the elements of ${\Omega}^{(j)}$ will be ordered based on the evaluations, where the elements with lowest maximum error rates ($\max$(${a}_p$, ${b}_p$)) are considered as the better solutions. Further, $n_q$ for the solutions that are dominated by each of optimal solutions will be reduced, and if $n_q=0$ then those solutions will be added to the set of Front solutions for the next step (${\Omega}^{(j+1)}$). This procedure will be continued in order to select the $O$ optimal solutions from $\mathbf{U}$.

\begin{figure}[!t]
	\begin{center}
		\includegraphics[width=0.48\textwidth,height=0.98\linewidth]{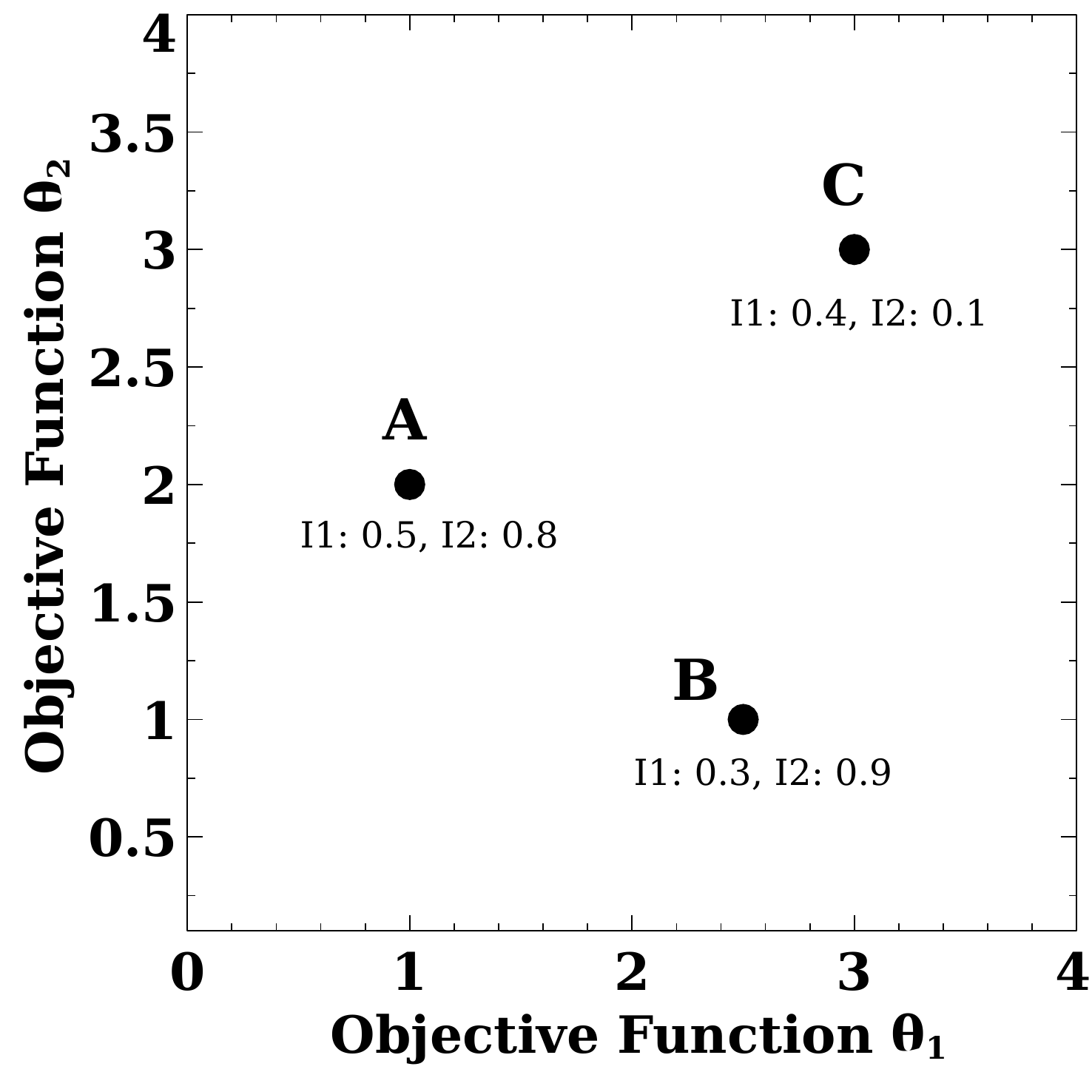}
		\caption{An example of selecting optimal solution by using MOCM.}
		\label{fig:Dominant}
	\end{center}
	\vskip -0.2in
\end{figure}

Figure \ref{fig:Dominant} shows an example of MOCM solution, where two objective functions generate three different solutions (i.e., A, B, and C). The solutions A and B can dominate the solution C, including both $\theta_1(C)$ and $\theta_2(C)$ are greater than other solutions. However, we cannot select A or B based on non-dominated sorting because of $\theta_1(A) < \theta_1(B)$ and $\theta_2(A) > \theta_2(B)$. Therefore, the indicators $\mathbf{I1}$ and $\mathbf{I2}$ are employed to evaluate the solutions A and B. Here, A is selected as the optimal solution, where the maximum of indicator values ($\max(\mathbf{I1}_A, \mathbf{I2}_A) = 0.8$) in solution A is lower than B ($\max(\mathbf{I1}_B, \mathbf{I2}_B) = 0.9$).

\begin{table*}[t]
	\caption{fMRI datasets}
	\label{tbl:Datasets}
	\begin{small}
		\begin{center}	
			\begin{tabular}{clccccccccccc}
				\hline
				ID & Title & X & Y & Z & \# & R & L & T & V & TR & TE &  Scanner  \\
				\hline
				DS005 &Mixed-gambles task & 53 & 63 & 52 &16 & 48 & 2 & 240 & 450 & 2 & 30& Siemens 3 Tesla  \\
				DS105  &Visual Object Recognition & 79 & 95 & 79 & 6 &71 & 8 & 121 & 1963 & 2.5 & 30 &  GE 3 Tesla \\
				DS107  &Word \& Object Recognition& 53 & 63 & 52 & 49 & 98 & 4 & 164 & 932 & 2 & 28& Siemens 3 Tesla  \\
				DS116  &Auditory and Visual Oddball & 53 & 63 & 40 & 17 &102 & 2 & 170 & 2532 & 2 & 25&  Philips 3 Tesla  \\
				DS117  & Multi-subject, multi-modal  & 64 & 61 & 33 & 19 &171 & 2 & 210 & 524 & 2& 30& Siemens 3 Tesla  \\
				CMU    &Meanings of Nouns & 51&61&23& 9 &9&12&360&17326&1&30&Siemens 3 Tesla\\
				\hline
			\end{tabular}
		\end{center}
		X, Y, Z as the size of 3D images;  $\# = S + \widehat{S}$ denotes the number of subjects; R is the number of all runs (sessions); L denotes the number of stimulus categories; T is the number of time points; V denotes the number of voxels in ROI;  TR is Time of Repetition in second; TE denotes Echo Time in millisecond.
	\end{small}
\end{table*}
\vskip -0.2in
\section{Experiments}
\subsection{Datasets}
This paper utilizes $6$ datasets, mostly shared by Open fMRI\footnote{Available at http://openfmri.org}, for running empirical studies in this paper. These datasets are listed as follows:
\begin{itemize}
	\item \textbf{DS005} includes $2$ categories of risk tasks with the $50/50$ chance of selection. In addition, Regions of Interest (ROI) is defined based on the original paper \cite{tom07}. 
	\item \textbf{DS105} includes $8$ categories of visual stimuli, i.e., gray-scale photos of cats, faces, houses, shoes, bottles, scissors, chairs, and scrambles (nonsense patterns). The neural activities in Ventral Temporal (VT) cortex is considered as the ROI in this dataset. Please refer to \cite{haxby11,haxby14} for technical information. 
	\item \textbf{DS107} contains 4 categories of visual stimuli, i.e., consonants, scrambles, objects, and words. The ROI is also defined based on the original study  \cite{duncan09}. 
	\item \textbf{DS117} includes MEG and fMRI images, where this paper just utilizes the fMRI data for running the empirical studies. Further, this dataset contains $2$ categories of visual stimuli, i.e., human faces, and scrambles. In this dataset, the voxel responses in the VT cortex are considered as the ROI. Please see \cite{wakeman15} for more information. 
	\item \textbf{DS116} contains EEG signals and fMRI images. We just use the fMRI data in order to generate the experiments. This data includes $2$ categories of audio and visual stimuli, including oddball tasks. Also, ROI is selected based on the original paper \cite{walz13}. 
	\item \textbf{CMU} includes 12 semantic categories of word photos as the visual stimuli. Here, the ROI is defined based on the intersection of coordinates across subjects. Please refer to \cite{mitchell08} for more information.
\end{itemize}
Table \ref{tbl:Datasets} summarizes the technical information of these datasets. Further, this paper separately preprocessed all datasets by using FSL 5.0.9\footnote{Available at https://fsl.fmrib.ox.ac.uk}, i.e., slice timing, anatomical alignment, normalization, smoothing. Here, we have utilized the standard HRF signal generated by FSL in order to convolve the task events.

\begin{table*}[!t]
	\caption{Accuracy of Classification Methods (mean$\pm$std)}
	\label{tbl:BinaryAccuracy}
	\begin{center}\begin{small}\begin{tabular}{lcccccc}
				\hline
				$\downarrow$Algorithms, Datasets$\rightarrow$ & DS005 & DS105 & DS107 & DS116 & DS117 & CMU \\
				\hline
				L1 SVM \cite{mohr15,bradley98}  & 71.65$\pm$4.97 &85.29$\pm$3.49& 81.25$\pm$3.62 & 69.24$\pm$3.28 & 76.61$\pm$2.73 & 73.62$\pm$3.15\\
				L1 SVM + HA \cite{haxby11,guntupalli16} & 81.27$\pm$3.59 & 87.03$\pm$2.87  & 84.01$\pm$1.56 & 74.62$\pm$1.84 & 77.93$\pm$2.29 & 80.23$\pm$1.63 \\
				L1 SVM + KHA \cite{lorbert12} & 83.06$\pm$2.36 & 90.05$\pm$2.39 & 86.68$\pm$1.71 & 80.51$\pm$2.12 & 84.22$\pm$1.44 & 83.49$\pm$2.03\\
				Osher et al. \cite{osher15}  & 84.55$\pm$2.02  & 90.82$\pm$1.87 & 85.62$\pm$1.95 & 78.91$\pm$2.04 & 86.81$\pm$1.79 & 85.01$\pm$1.97 \\
				PSO-SVM \cite{ma16} & 70.32$\pm$1.92 & 77.91$\pm$1.03 & 81.21$\pm$2.33 & 76.14$\pm$1.49 & 83.71$\pm$2.81 & 82.61$\pm$1.05 \\
				HHPSO-SVM \cite{ma16} & 90.17$\pm$1.01 & 94.46$\pm$1.23 & 89.91$\pm$1.67 & 96.03$\pm$0.56 & \textbf{96.74$\pm$1.01} &87.62$\pm$1.03\\
				Kao et al. \cite{kao12} & 89.68$\pm$0.87 & 95.31$\pm$0.44 & 87.77$\pm$0.28 & 84.16$\pm$0.73 & 90.49$\pm$0.39 & 90.93$\pm$1.82\\
				MOMVP (Linear kernel) & 94.79$\pm$0.57 & 93.61$\pm$0.57 & 92.83$\pm$0.57 & 90.93$\pm$0.71 & 91.90$\pm$0.27 & 94.37$\pm$0.98\\
				MOMVP (Gaussian kernel)& \textbf{96.10$\pm$0.29} & \textbf{98.34$\pm$0.29} & \textbf{96.79$\pm$0.59} & \textbf{97.09$\pm$0.33} & 95.49$\pm$0.18 & \textbf{96.02$\pm$0.92}\\	
				\hline
	\end{tabular}\end{small}\end{center}
	\vskip -0.1in
\end{table*}
\begin{table*}[!h]
	\caption{Area Under the ROC Curve (AUC) of Classification Methods (mean$\pm$std)}
	\label{tbl:BinaryAUC}
	\begin{center}\begin{small}\begin{tabular}{lcccccc}
				\hline
				$\downarrow$Algorithms, Datasets$\rightarrow$ & DS005 & DS105 & DS107 & DS116 & DS117 & CMU \\
				\hline
				L1 SVM \cite{mohr15,bradley98} & 68.37$\pm$4.01 & 80.91$\pm$3.21 & 80.72$\pm$2.88 & 66.85$\pm$3.05 & 72.12$\pm$1.48 & 70.08$\pm$2.94\\
				L1 SVM + HA \cite{haxby11,guntupalli16} & 70.32$\pm$2.92 & 84.82$\pm$2.53 & 82.94$\pm$1.03 & 73.91$\pm$1.33 & 75.14$\pm$2.49 & 78.32$\pm$1.21\\
				L1 SVM + KHA \cite{lorbert12} & 82.22$\pm$2.42 & 88.81$\pm$1.61 & 83.36$\pm$2.12 & 77.41$\pm$1.97 &  81.54$\pm$1.92  & 81.76$\pm$2.92\\
				Osher et al. \cite{osher15} & 81.83$\pm$2.86 & 89.54$\pm$1.74 & 82.02$\pm$2.43 & 75.08$\pm$1.12 & 84.08$\pm$1.84 & 82.27$\pm$2.06\\
				PSO-SVM \cite{ma16} & 67.84$\pm$2.82 & 75.61$\pm$1.57 & 80.14$\pm$2.47 & 73.59$\pm$1.95 & 79.05$\pm$2.12 & 79.88$\pm$3.73\\
				HHPSO-SVM \cite{ma16} & 87.91$\pm$1.83 & 92.39$\pm$1.73 & 86.12$\pm$0.99 & \textbf{95.32$\pm$1.18} & 90.73$\pm$1.59 & 87.01$\pm$1.61\\
				Kao et al. \cite{kao12} & 88.13$\pm$1.58 & 90.45$\pm$0.73 & 84.67$\pm$1.04 & 83.28$\pm$1.47 & 89.69$\pm$1.27 & 89.00$\pm$1.02\\
				MOMVP (Linear kernel) & 91.17$\pm$0.80 & 92.36$\pm$0.84 & 91.63$\pm$0.69 & 90.16$\pm$0.72 & 90.15$\pm$0.69 & 92.66$\pm$0.73\\
				MOMVP (Gaussian kernel) & \textbf{94.37$\pm$0.63} & \textbf{97.71$\pm$0.58} & \textbf{93.22$\pm$0.49} & 94.97$\pm$0.14 & \textbf{93.72$\pm$0.31} & \textbf{95.79$\pm$0.42} \\
				\hline
	\end{tabular}\end{small}\end{center}
	\vskip -0.2in
\end{table*}

\subsection{Performance Analysis}
This section compares the performance of the proposed method with different MVP techniques. As a baseline, this paper reports the performance of L1 SVM \cite{bradley98}, which is used in \cite{mohr15} as the best algorithm for MVP analysis. Further, the performance of the original HA \cite{guntupalli16,haxby11} and KHA \cite{lorbert12} are addressed for demonstrating the effect of functional alignment in MVP analysis. Here, KHA algorithm is applied by using the Gaussian kernel that introduced as the best kernel in the original paper \cite{lorbert12}. Further, we utilized $\sfrac{1}{n}$ as the gamma parameter for all employed Gaussian kernels in this paper, where $n$ is the number of features \cite{lorbert12,tony17b}. As a new graph-based method, the performance of Osher et al. method \cite{osher15} is also reported in this paper. As a baseline for single-objective swarm optimization techniques in MVP analysis, the performance of PSO-SVM \cite{ma16} is addressed in this paper. Moreover, the performance of HHPSO-SVM \cite{ma16} and Kao et al. method \cite{kao12} are reported as two multi-objective-based methods in MVP analysis. Finally, the performance of the MOCM method is addressed by using two different mapping functions, i.e., a linear mapping (${\Phi}(\mathbf{x})=\mathbf{x}$), and Gaussian kernel same as KHA method. In addition, there is no feature selection in this section. Like \cite{ma16}, the population size is considered $O=50$ for PSO, HHPSO, NSGA-II (in Kao et al. method), and MOCM. Moreover, we consider $MaxIt=1000$ and $MaxSame=5$ for all datasets. Like the previous studies \cite{lorbert12,mohr15,tony16,tony17a,tony17b}, this paper firstly partitions the original fMRI images to training-set and testing-set by using leave-one-subject-out cross-validation. Then, the general linear model is calculated in the subject-level for all methods. After that, the functional alignment parameters are calculated for hyperalignment techniques. Next, this paper generates binary classifiers by applying one-versus-all strategy to the neural activities in the training sets. Finally, the performances of trained classifiers are evaluated by applying unseen testing sets and calculating the average of accuracy (or AUC) \cite{tony16,tony17b}. It is worth noting that the same structure and sample sets are applied to all evaluated methods in each iteration. Furthermore, the mentioned algorithms are implemented in the MATLAB R2016b (9.1) on a PC with certain specifications\footnote{DEL, CPU = Intel Xeon E5-2630 v3 (8$\times$2.4 GHz), RAM = 64GB, OS = Ubuntu 16.04.2 LTS} by authors for generating the empirical studies.

Table \ref{tbl:BinaryAccuracy} and \ref{tbl:BinaryAUC} respectively illustrate the classification Accuracy and Area Under the ROC Curve (AUC) in percentage (\%). As depicted in these tables, L1 SVM cannot provide acceptable performance in comparison with other techniques because it just uses the anatomical alignment. Further, functional alignment techniques (HA and KHA) improved the performance of MVP analysis in comparison with L1 SVM method. In addition, HHPSO-SVM generated better results in comparison with PSO-SVM because it uses a multi-objective optimization approach. Moreover, the performance of Kao et al. method is significantly unstable because the optimization approach (NSGA-II) in this method cannot trace errors very well. Indeed, this is the main reason that we extend the indicators algorithm for improving the robustness of non-dominated sorting. Here, the performances of multi-objective approaches are more stable than the singular-objective methods (based on the standard deviation). Finally, the proposed method has generated better performance in comparison with other methods because it provided a robust and stable solution for MVP analysis by developing an integrated objective function and providing an effective optimization strategy. Indeed, the proposed method provides better performance when it is applied by using the Gaussian kernel that can map the nonlinear data points to a linear space. Furthermore, MOCM can calibrate the parameters generated in each step of fMRI analysis by tracing the errors in other steps. A good example is functional alignment techniques that can generate different solutions for a specific problem \cite{chen15}. While a single objective function generates these solutions, there is no way to rank or select one of them. However, MOCM can rank all possible solutions in each step of fMRI analysis (i.e., function aligning, classification, etc.) by tracing the effects of that solution on the other steps. Here, if we have two different alignment solutions for a specific problem, MOCM selects the solution with lowest classification error as the optimal solution. It is worth noting that we always select a set of optimal solutions in each iteration that has potential to generate better solutions in the next step (by creating new offsprings).

\subsection{MVP analysis by using feature selection }
\begin{figure}[t]
	\begin{center}
		\begin{minipage}{0.48\linewidth}
			\includegraphics[width=0.99\textwidth,height=0.9\linewidth]{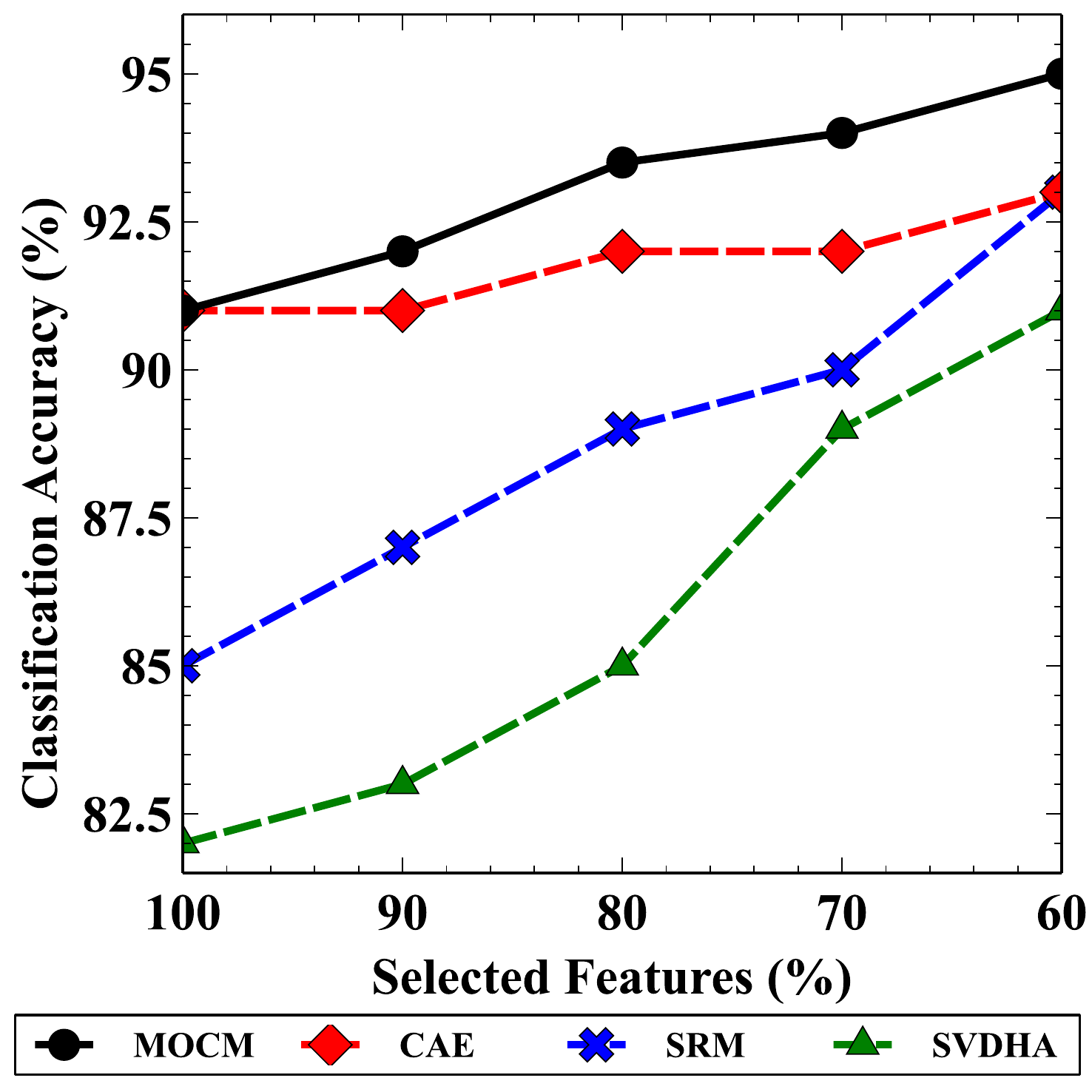}\\	
			\centering (A) DS005\\
		\end{minipage}
		\begin{minipage}{0.48\linewidth}
			\includegraphics[width=0.99\textwidth,height=0.9\linewidth]{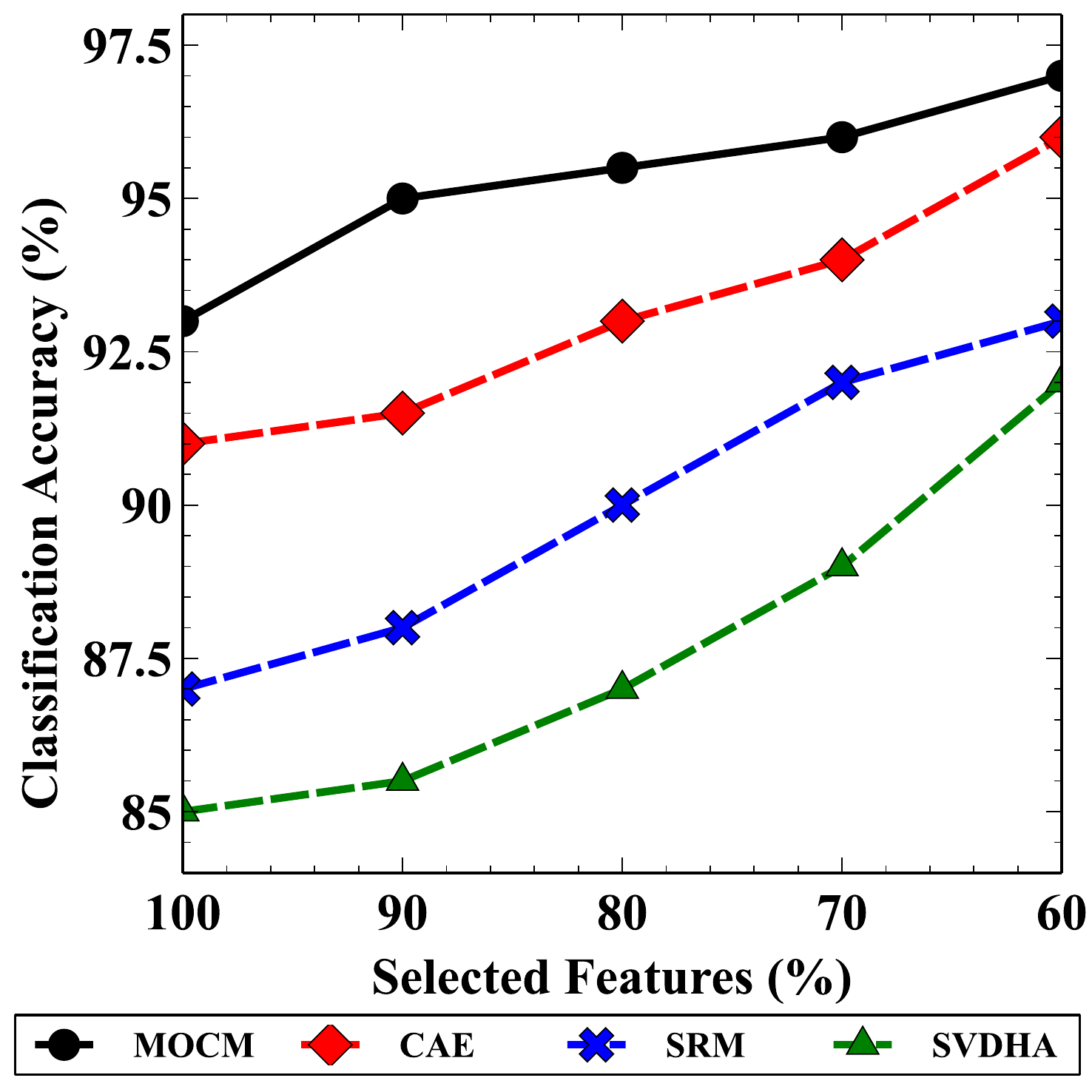}\\	
			\centering (B) DS105
		\end{minipage}
		\begin{minipage}{0.48\linewidth}
			\includegraphics[width=0.99\textwidth,height=0.9\linewidth]{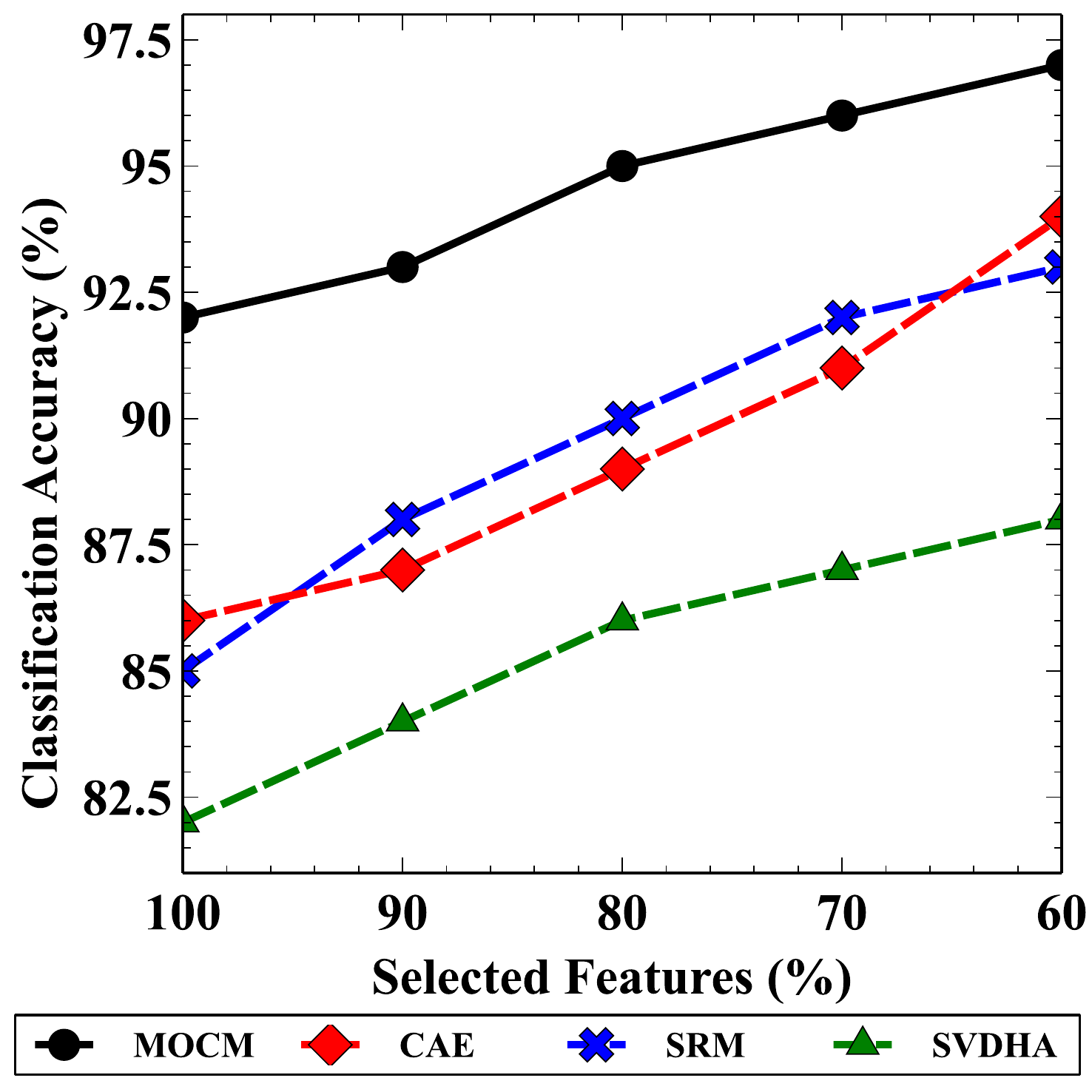}\\	
			\centering (C) DS107
		\end{minipage}
		\begin{minipage}{0.48\linewidth}
			\includegraphics[width=0.99\textwidth,height=0.9\linewidth]{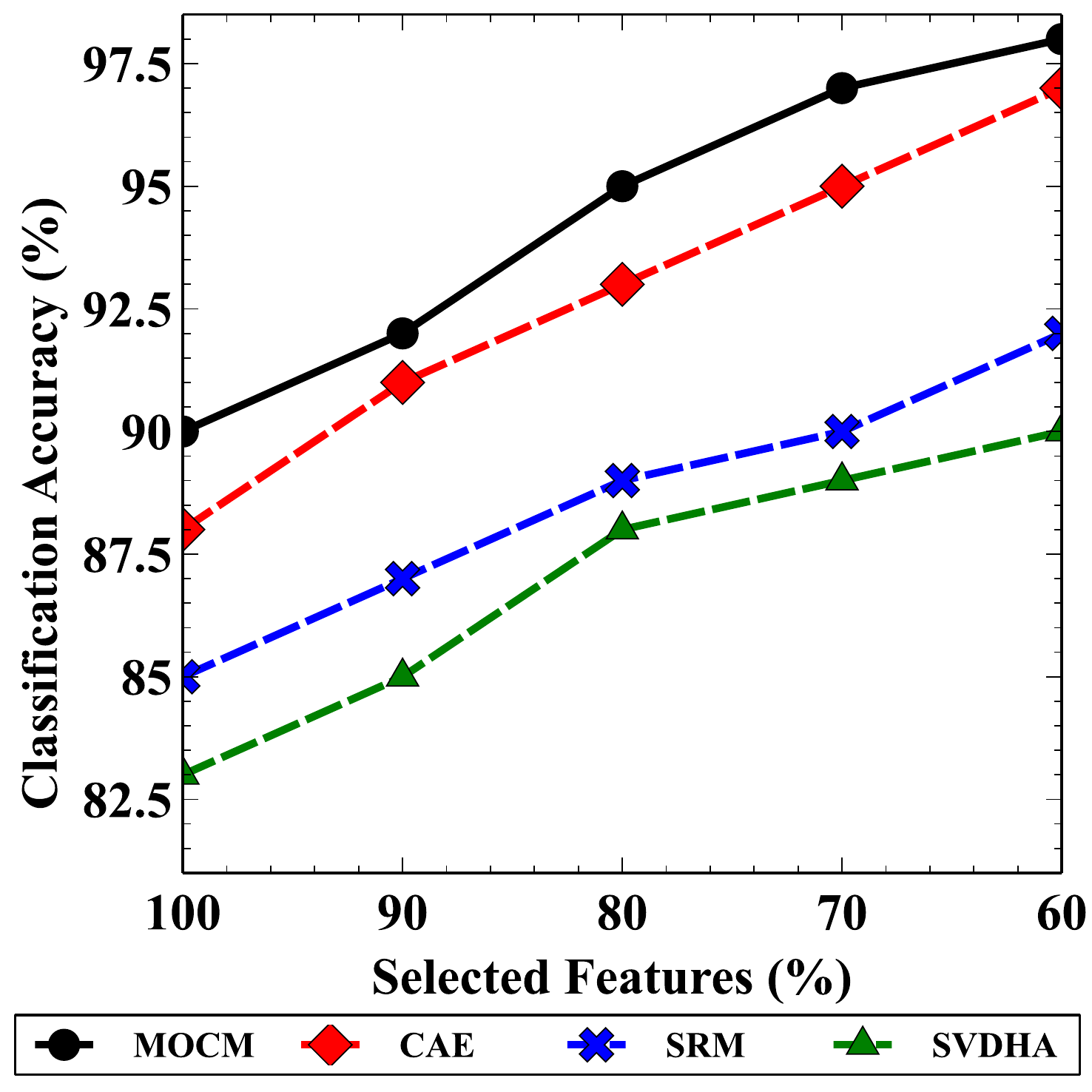}\\	
			\centering (D) DS116
		\end{minipage}
		\begin{minipage}{0.48\linewidth}
			\includegraphics[width=0.99\textwidth,height=0.9\linewidth]{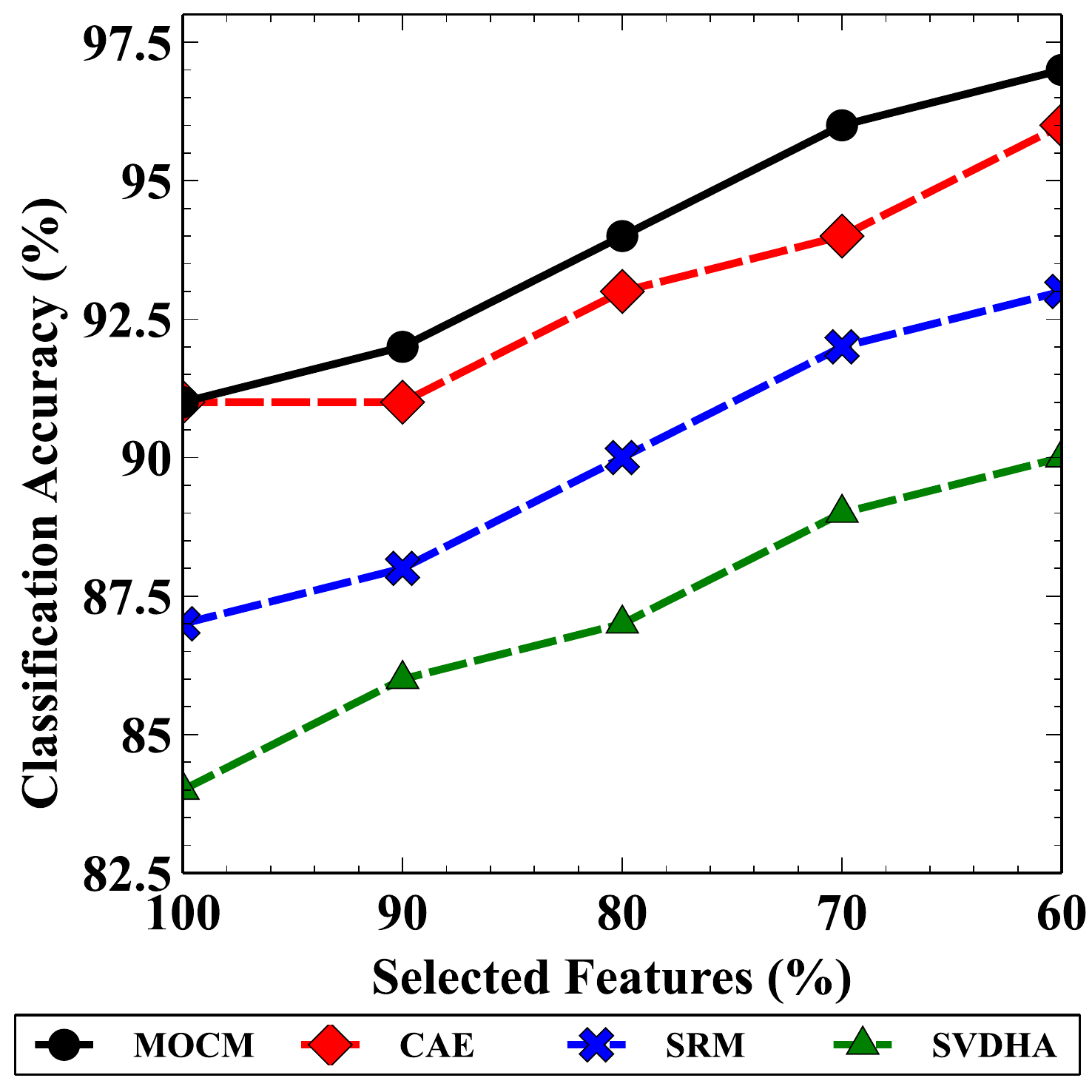}\\	
			\centering (E) DS117
		\end{minipage}
		\begin{minipage}{0.48\linewidth}
			\includegraphics[width=0.99\textwidth,height=0.9\linewidth]{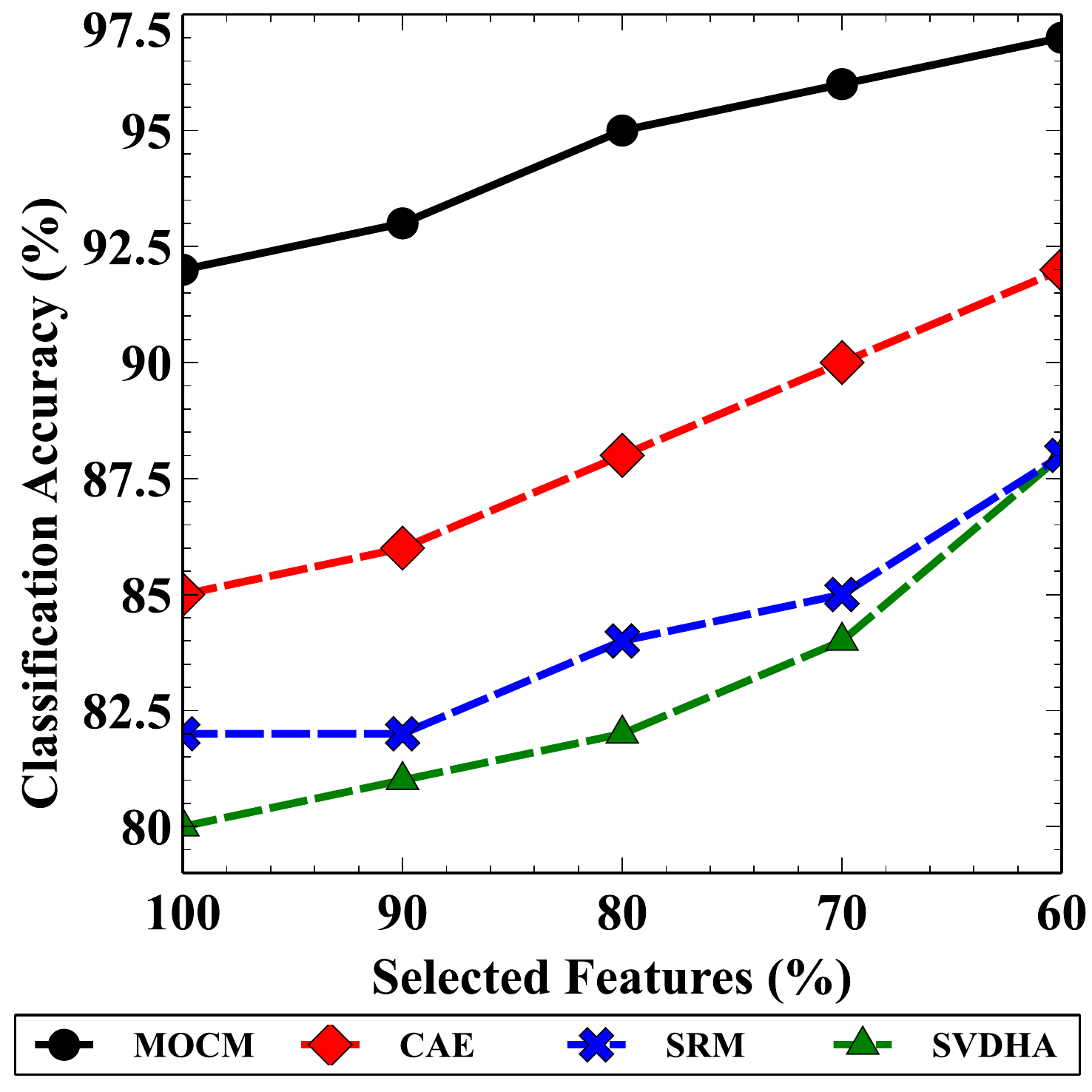}\\	
			\centering (F) CMU
		\end{minipage}	
		\caption{MVP analysis by using feature selection.}
		\label{fig:FeatureSelection}
	\end{center}
	\vskip -0.25in
\end{figure}

This section analyzes the performance of MVP methods by using the features selection techniques. MOCM is compared with SVDHA \cite{chen14}, SRM \cite{chen15}, and CAE \cite{chen16} as the state-of-the-art MVP methods that can apply feature selection before generating a cognitive model. Here, L1 SVM is used for generating the cognitive models after each of the mentioned methods are applied on the preprocessed fMRI images for functional alignment. Like SVDHA, the proposed method employs a feature section function in terms of SVD analysis, where the mapping function ${\Phi}:V_{org}\to V, V_{org} \gg V$ is defined in order to generate the cognitive model \cite{chen14}. In other words, SVD decomposition is applied to the neural activities and then features are sorted based on the largest singular values. After that, we have selected the $V$ features, where they have the largest $V$ singular values in the decomposition. Next, the selected features are used in SVDHA and MOCM for training the classification model. For CAE method, the features are selected by reducing the number of units in the convolution neural network. In addition, we have selected features in SRM by changing the parameter $k$ in this method, where $k$ is the size of features for generating the mappings ($\mathbf{W}_i$) and the shared space ($\mathbf{S}$) in SRM method \cite{chen15}. It is worth noting that the feature selection procedure is applied separately to the training set and the testing set after these sets are partitioned by using cross-validation. Further, the setup of this experiment is same as the previous section (cross-validation, the population size, etc.). Figure \ref{fig:FeatureSelection} shows the performance of different methods by selecting $100\%$ to $60\%$ of features. As shown in this figure, the proposed method has generated better performance in comparison with other methods. Indeed, it can track errors of learning during the feature selection and then update different training coefficients ($\beta^{(i)}$, $\mathbf{R}^{(i)}$, $\mathbf{W}$) for minimizing the generated errors. 
\subsection{Runtime Analysis}
\begin{figure*}[!t]
	\centering
	\begin{minipage}[b]{0.32\linewidth}
		\includegraphics[width=0.99\textwidth,height=0.62\linewidth]{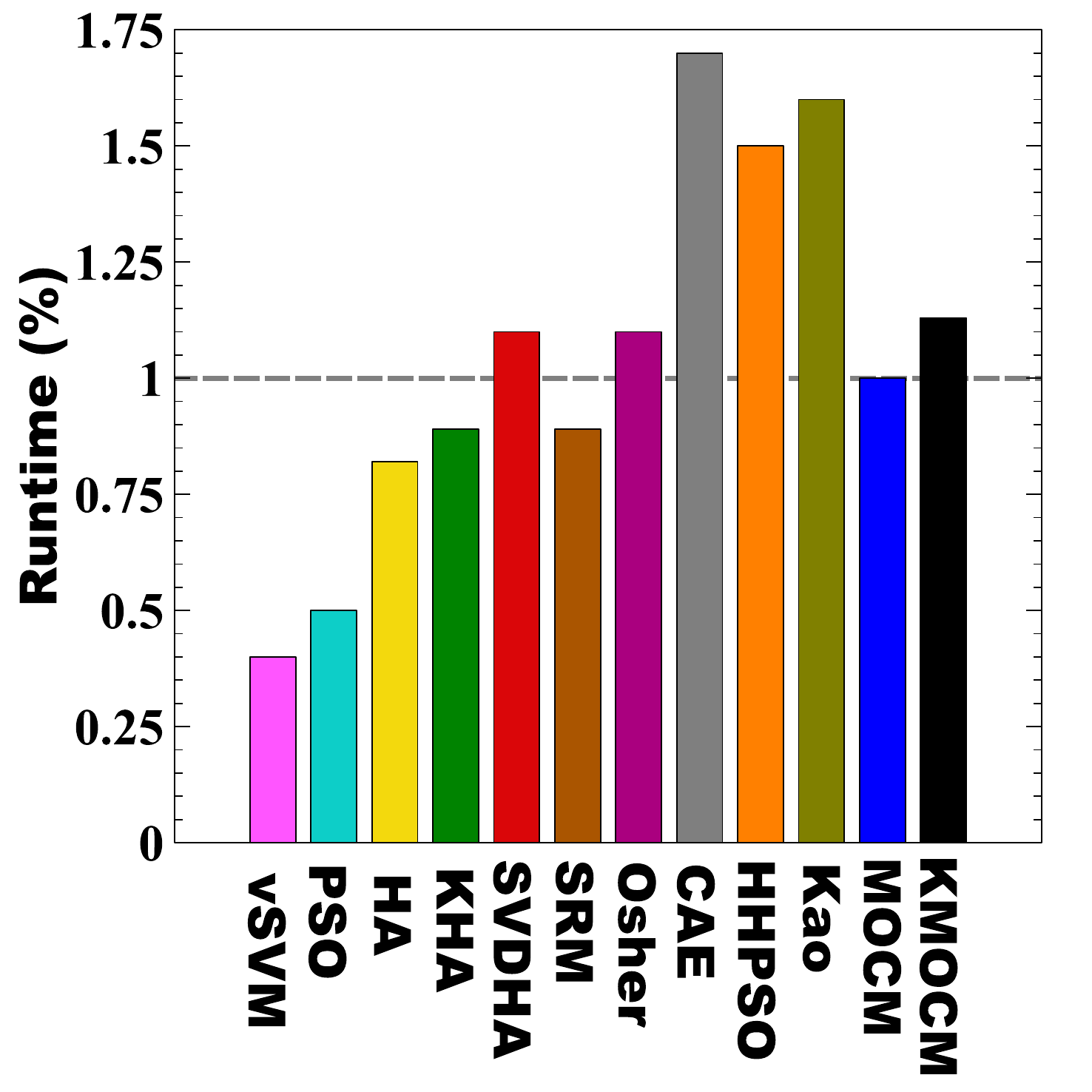}
		\\	\centering {\small (A) DS005}
	\end{minipage}
	\begin{minipage}[b]{0.32\linewidth}
		\includegraphics[width=0.99\textwidth,height=0.62\linewidth]{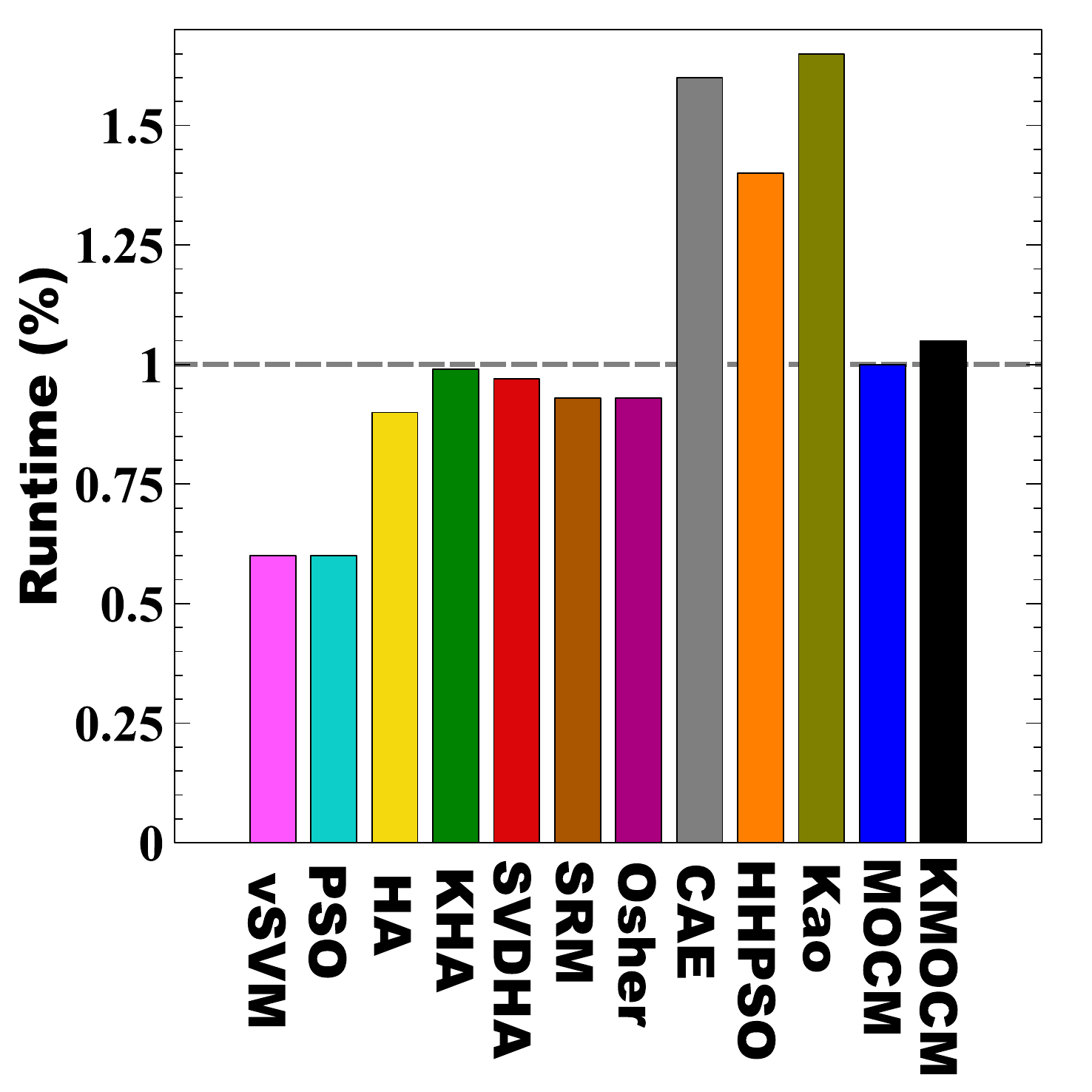}
		\\	\centering {\small (B) DS105}
	\end{minipage}
	\begin{minipage}[b]{0.32\linewidth}
		\includegraphics[width=0.99\textwidth,height=0.62\linewidth]{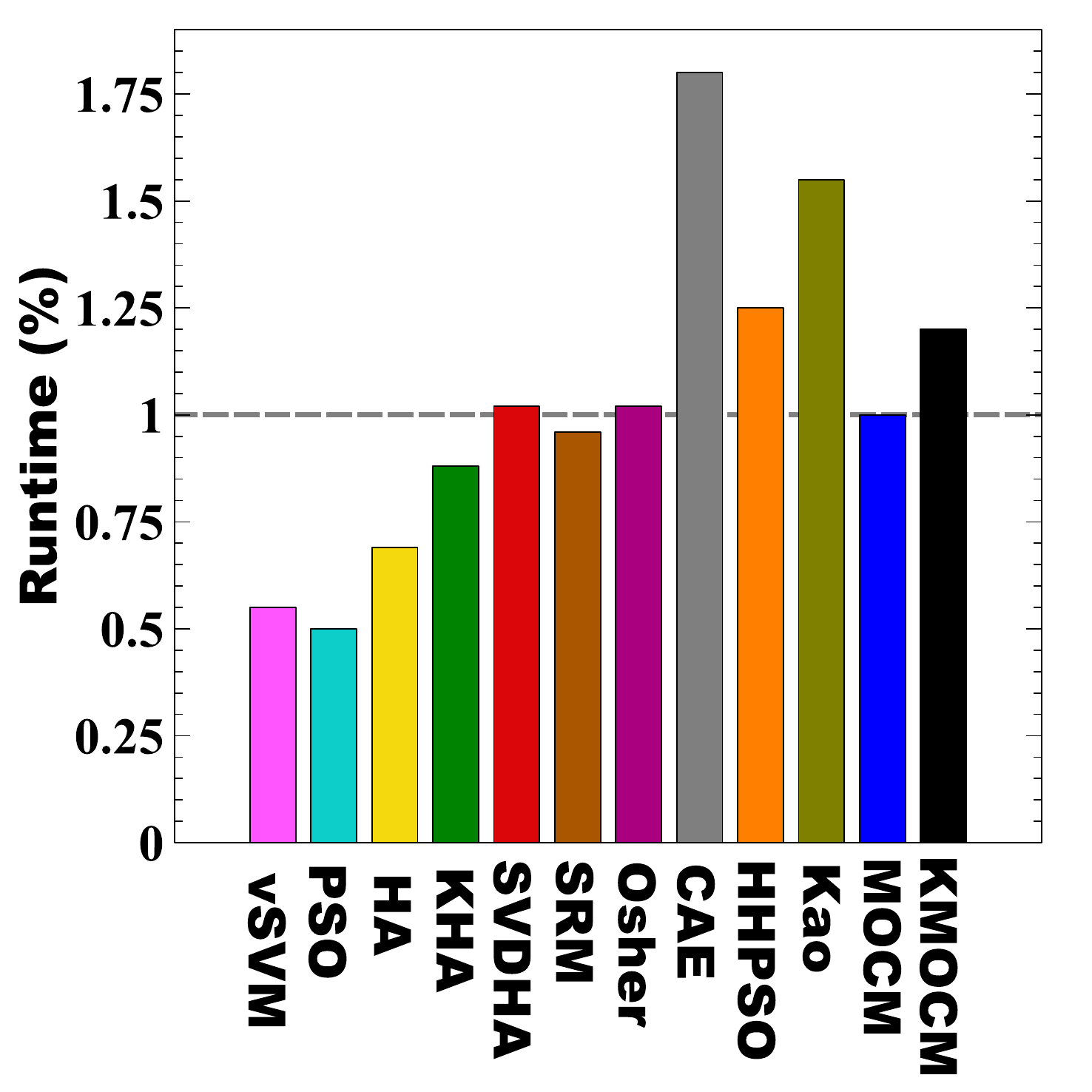}
		\\	\centering {\small (C) DS107}
	\end{minipage}
	\begin{minipage}[b]{0.32\linewidth}
		\includegraphics[width=0.99\textwidth,height=0.62\linewidth]{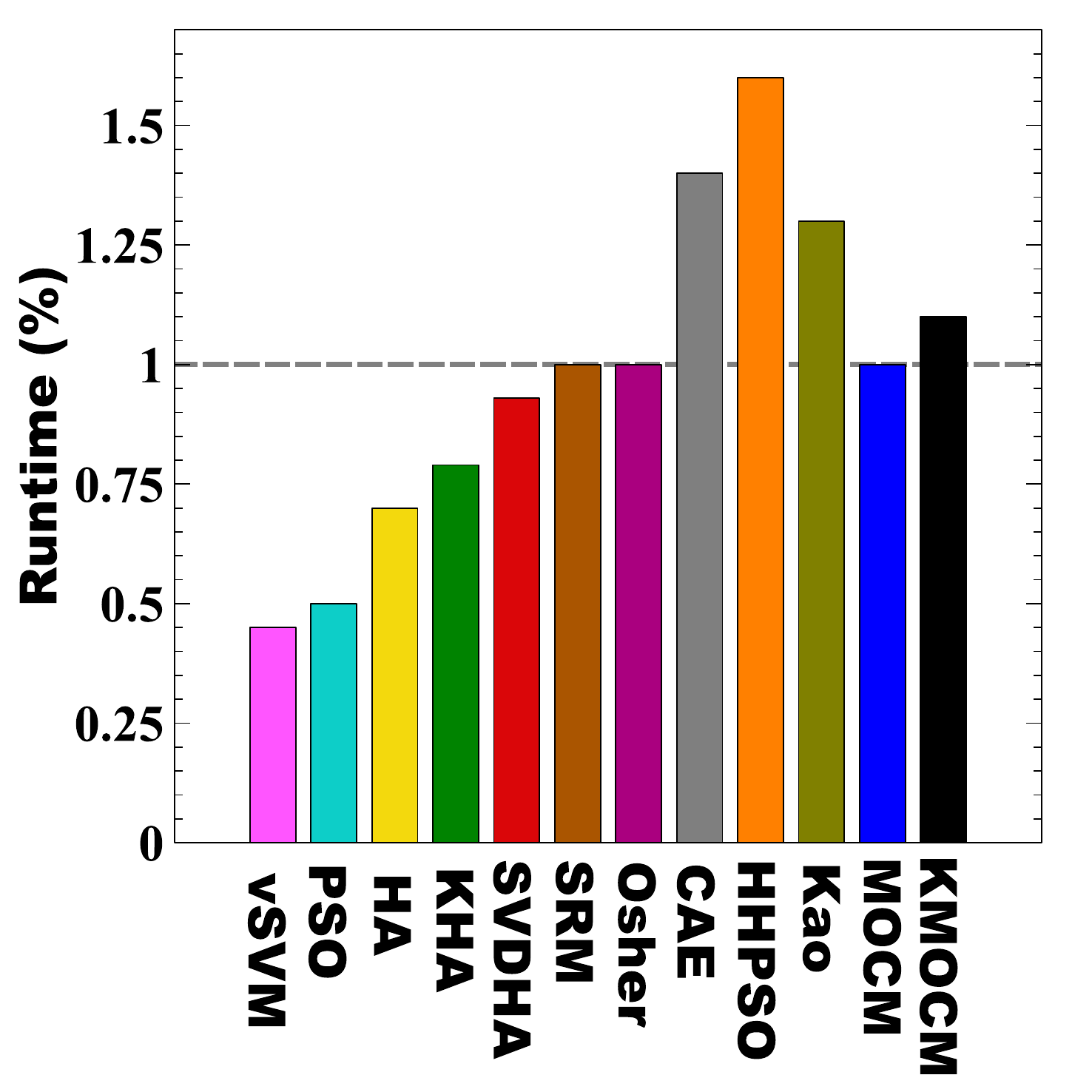}
		\\	\centering {\small (D) DS116}
	\end{minipage}
	\begin{minipage}[b]{0.32\linewidth}
		\includegraphics[width=0.99\textwidth,height=0.62\linewidth]{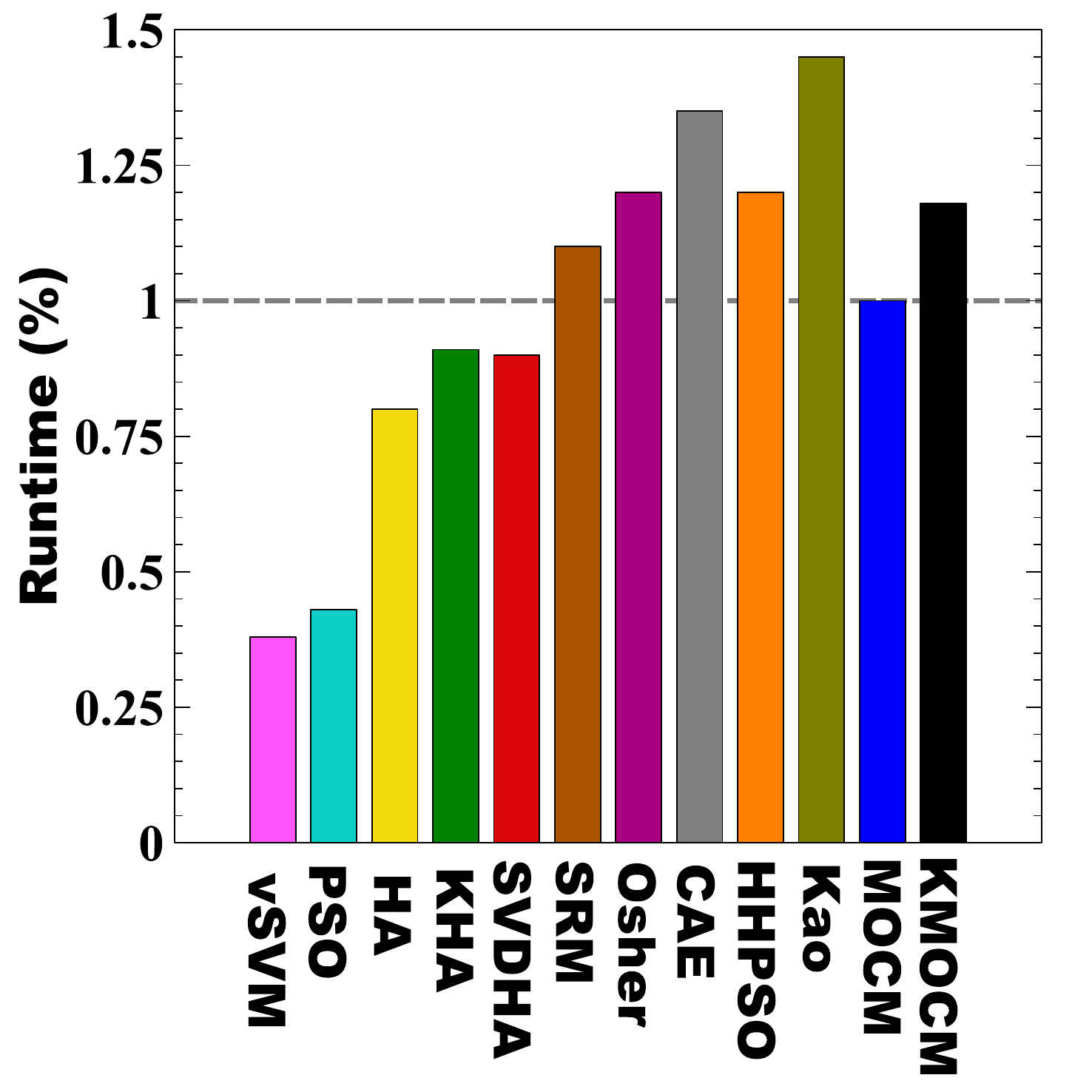}
		\\	\centering {\small (E) DS117}
	\end{minipage}
	\begin{minipage}[b]{0.32\linewidth}
		\includegraphics[width=0.99\textwidth,height=0.62\linewidth]{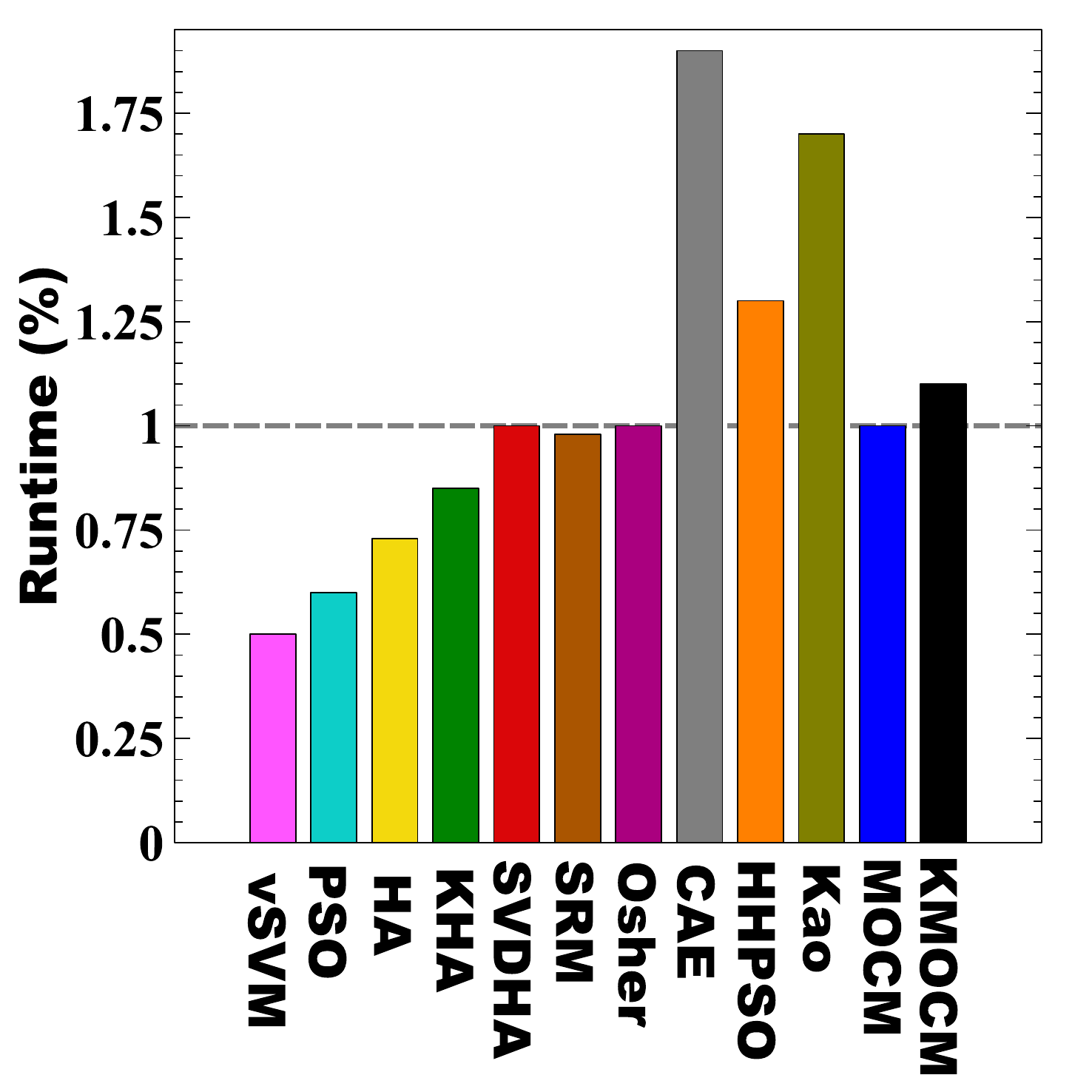}
		\\	\centering {\small (F) CMU}
	\end{minipage}
	\caption{Runtime Analysis}
	\label{fig:Runtime}
	\vskip -0.2in        
\end{figure*}

This section analyzes runtime of different MVP methods. As mentioned before, all of the empirical studies are generated by using a specified PC. Figure \ref{fig:Runtime} compares the runtime of MOCM with other functional alignment methods, where all runtime are scaled based on the proposed method (the runtime of MOCM is utilized as a unit). As this figure illustrates, there are four groups of methods based on the runtime. As the first group, SVM \cite{mohr15,bradley98} and PSO-SVM \cite{ma16} just employed a singular objective function and data without functional alignment. Therefore, they produce low accuracy (see previous sections) and runtime. As the second group, HA \cite{haxby11}, KHA \cite{lorbert12}, SVDHA \cite{chen14}, Osher et al. method \cite{osher15}, SRM \cite{chen15}, and CAE \cite{chen16} simultaneously utilized two singular objective functions for functional alignment and classification learning. Since HA, KHA, and SVDHA employed a single objective function for generating the function alignment parameters, i.e., the shared space ($\mathbf{G}$) and mapping functions ($\mathbf{R}^{(i)}$), the number of iteration for optimizing these parameters is naturally lower than a multi-objective solution. Thus, they are a little faster than MOCM. However, the performance of these methods are limited, and the optimization approaches in these methods cannot calibrate the alignment parameters based on the generated errors in GLM step or classification learning procedure. It is worth noting that the runtime of CAE is high because it employs deep learning method for aligning the neural activities. As the next group, HHPSO \cite{ma16} and Kao et al. \cite{kao12} methods use the multi-objective approaches but just for the learning step. Indeed, these methods considered the functional alignment as the preprocessing step. By contrast, the proposed method does not need a separate step for functional alignment because it utilizes an integrated solution in order to apply the whole of procedures.

\begin{figure}[!t]
	\centering
	\includegraphics[width=0.48\textwidth]{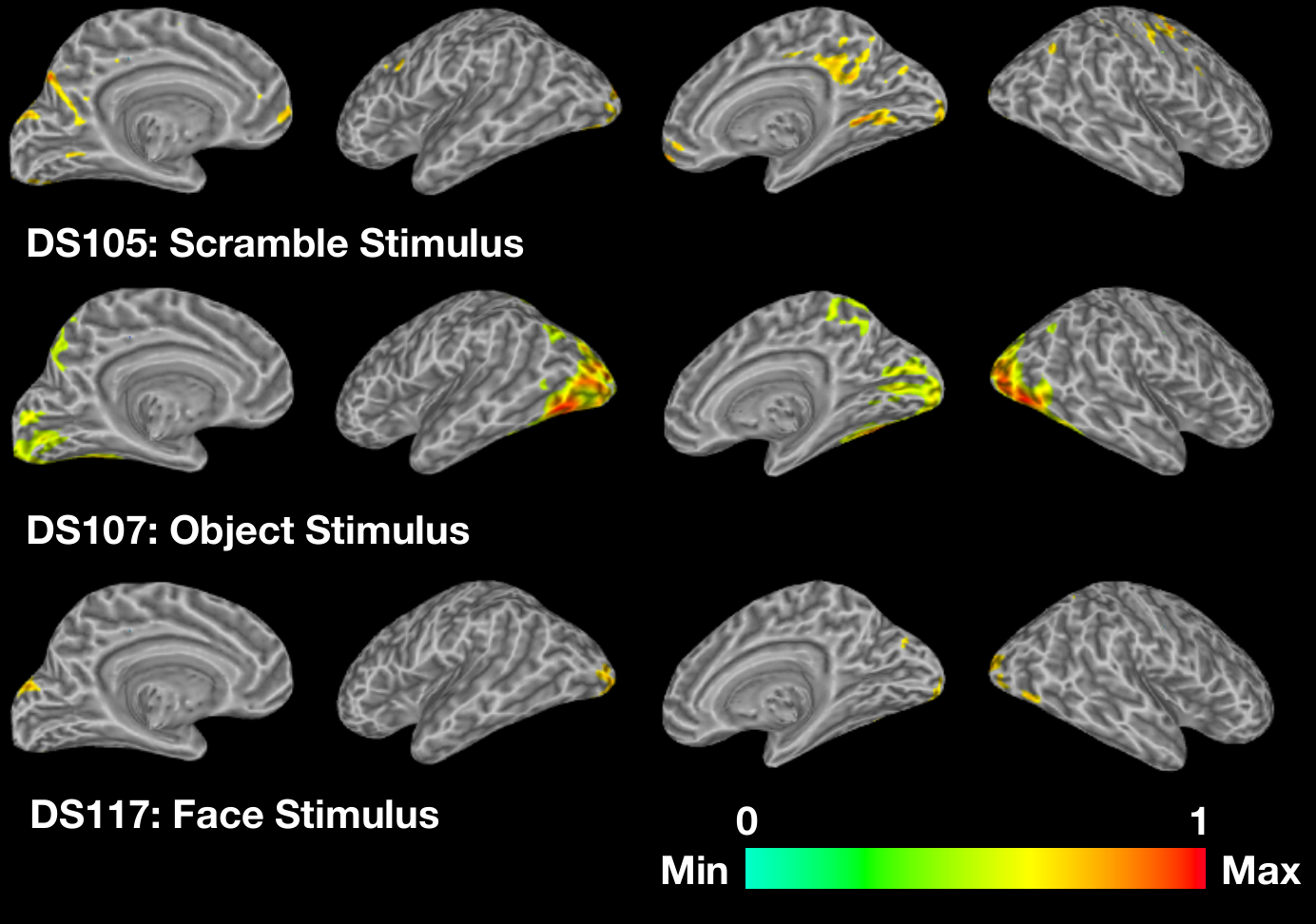}
	\caption{Examples of cognitive models (decision surfaces), generated for each category of stimuli by applying MOCM to whole-brain datasets}
	\label{fig:BrainActivities}
	\vskip -0.2in        
\end{figure}  

\section{Discussions and Conclusions}
As the final product of Multi-Objective Cognitive Model (MOCM), Figure \ref{fig:BrainActivities} depicts some examples of the generated cognitive models across categories of stimuli. Indeed, we visualized the decision surfaces ($\mathbf{W}$) that are generated in the training-phase. In order to create the cognitive model, we applied MOCM with a linear mapping (${\Phi}(\mathbf{x})=\mathbf{x}$) to the whole-brain fMRI images with the following parameters: $O=50, MaxIt=1000, MaxSame=10$. This figure illustrates that different loci are activated based on distinctive stimuli. Further, the brain activities will be more focused in a certain region, when the stimuli just include the specific exemplars (such as human faces in DS117) rather than the abstract categories (concepts), e.g. objects in DS107. Indeed, this assumption is matched by the results of the previous studies \cite{haxby11,mohr15,mitchell08,tony16,tony17b}, and can be considered as a shred of evidence for validating the generated model. It is worth noting that the proposed method can used for understanding how the human brain works and seeking new treatments for mental diseases.

There are several advantages to using multi-objective approach. Firstly, it can simplify the procedure of analysis. While other approaches need different steps with distinctive parameters (that may conflict with each other), we only need to apply a single step in the MOCM method for generating every thing, i.e., beta values, aligned features, and the classification model. The second advantage is tracing errors in different steps. Since we optimize a vector (i.e., the cost of different objective functions) at the same time rather than the disjoint single objective functions, we can rank and select possible solutions based on their generated errors in different steps. Since the set of optimal solutions (new offsprings) in each iteration are generated by using the ranked solutions of the previous iteration, they have potential to improve the quality of the final results in all steps simultaneously, including beta values, aligned features, classification models, etc.

In summary, this paper proposes MOCM as an integrated objective function in order to improve the performance and stability in the supervised fMRI analysis. By contrast of the previous methods, this objective function can apply both the functional alignment step and the learning step at the same time. Further, this objective function is generalized by using the kernel approach (for nonlinear data) and feature selection technique (for reducing the sparsity and noise). In order to solve the integrated objective function, a customized multi-objective optimization approach is developed by incorporating the idea of non-dominated sorting into the multi-indicator algorithm. Indeed, non-dominated sorting seeks all possible solutions, and then indicators rank the robust solutions as the final results. Empirical studies on multi-subject fMRI datasets confirm that the proposed method achieves superior performance to other state-of-the-art MVP techniques. In the future, we will plan to utilize the proposed method for improving the performance of other techniques in fMRI analysis, i.e unsupervised learning in RSA methods, multi-modality analysis, and neural hub detection. 

\section*{Information Sharing Statement}
The publicly available Open fMRI datasets are used in this paper that can be found via the GitHub site link: \url{https://openfmri.org}. Further, we have shared a preprocessed version of datasets in MATLAB format at \url{https://easyfmridata.github.io}. In addition, the proposed method can be accessed by using our GUI-based toolbox, i.e., available at \url{https://easyfmri.github.io}.

\section*{Compliance with Ethical Standards}
\section*{Conflict of Interests }
Muhammad Yousefnezhad and Daoqiang Zhang declare that they have no conflict of interest.
\section*{Ethical Approval}
This article does not contain any studies with human participants or animals performed by any of the authors.
\begin{acknowledgements}
This work was supported in part by the National Natural Science Foundation of China (61422204 and 61473149), and NUAA Fundamental Research Funds (NE2013105).
\end{acknowledgements}


%
%

\bibliographystyle{spmpsci}      
\bibliography{NIF17}   

%
%

\end{document}